\documentclass{amia}

\usepackage{graphicx}

\usepackage[labelfont=bf]{caption}
\usepackage{color}

\usepackage[square,sort,numbers]{natbib}

\usepackage{amsmath}
\usepackage{amssymb}
\usepackage{amsfonts}
\usepackage{booktabs}
\usepackage{url}
\usepackage{multirow}
\usepackage{multicol}
\usepackage{bbm}

\makeatletter
\newcommand\notsotiny{\@setfontsize\notsotiny\@vipt\@viipt}
\makeatother

\usepackage{algorithm}
\usepackage{longtable}

\begin{document}

\title{Identifying Distinct, Effective Treatments for Acute Hypotension with SODA-RL: Safely Optimized Diverse Accurate Reinforcement Learning}

\author{Joseph Futoma, PhD$^{1,2}$, Muhammad Masood, PhD$^{1}$, Finale Doshi-Velez, PhD$^{1}$}

\institutes{
    $^1$ Harvard University, Paulson School of Engineering and Applied Sciences, Cambridge, MA \\
    $^2$ Duke University, Dept. of Statistical Science, Durham, NC \\
}

\maketitle

\noindent{\bf Abstract}

\textit{Hypotension in critical care settings is a life-threatening emergency that must be recognized and treated early. While fluid bolus therapy and vasopressors are common treatments, it is often unclear which interventions to give, in what amounts, and for how long.  Observational data in the form of electronic health records can provide a source for helping inform these choices from past events, but often it is not possible to identify a single best strategy from observational data alone.  In such situations, we argue it is important to expose the collection of plausible options to a provider.  To this end, we develop SODA-RL: Safely Optimized, Diverse, and Accurate Reinforcement Learning, to identify distinct treatment options that are supported in the data. We demonstrate SODA-RL on a  cohort of 10,142 ICU stays where hypotension presented.  Our learned policies perform comparably to the observed physician behaviors, while providing different, plausible alternatives for treatment decisions.}

\section*{Introduction}

Patients in the intensive care unit (ICU) are among the sickest in the hospital, and require many different types of interventions to control and respond to their unstable physiological conditions. For instance, antibiotics are given to control infections \citep{ibrahim2000influence}, and anticoagulants are given to dialysis patients to prevent thrombosis \citep{berbece2006sustained}.  Patients with the highest acuity may be given more aggressive and invasive interventions such as mechanical ventilation \citep{esteban2002characteristics} as well.  

In this paper, we focus on decisions to give fluid bolus therapy \citep{glassford2014physiological} and vasopressors \citep{havel2011vasopressors} when treating hypotension and shock.  Hypotension is associated with overall higher morbidity and mortality in across several populations, including populations with sepsis \citep{maheshwari2018relationship} and populations in the emergency department \citep{jones2006emergency}. However, despite the importance of addressing this problem, decision making for hypotension management is not standardized, and treating these patients effectively is challenging.  Although it has been studied extensively \citep{girkar2018predicting}, the choice of bolus size and timing, as well as which vasopressor to use and in what dosing regimen is not well understood.

Reinforcement learning (RL), a branch of machine learning focused on learning how to make a sequence of decisions toward some desired outcome \citep{sutton1998introduction}, has the potential to help us use past data to assist with these decisions.  Recent applications of RL to healthcare include managing sepsis \citep{komorowski2018artificial}, schizophrenia \citep{shortreed2011informing}, mechanical ventilation \citep{prasad2017reinforcement}, and heparin dosing \citep{nemati2016optimal}. However, as noted in \citep{gottesman2019guidelines}, quantifying the quality of a proposed treatment policy is challenging.  Observational data create hard limitations on the kinds of policies that can be credibly evaluated: one cannot evaluate policies that recommend treatments that were never or rarely performed, and even when the recommended treatments have support in the observed data, the value of different choices may be impossible to statistically differentiate.

Thus, instead of attempting to identify a single optimal treatment policy from observational data---which is often impossible---in this work, we focus on identifying a \emph{collection} of \emph{distinct, plausible} policies.  Having such a collection of options can provide insights into multiple versions of treatments that may be of similar efficacy, and it also provides a step toward providing personalized recommendations by creating a space of reasonable treatment options.  One way to think about this approach is to note that the variation that we see in clinician actions is likely to be safe---patients are typically treated conservatively to avoid iatrogenic harm.  Amid this variation, our goal is to identify a collection of treatment policies that are both distinct---that is, different from each other, so as to provide choices of options---but also likely---that is, are not too far from current practices. To this end, in this work we develop SODA-RL: Safely Optimized, Diverse, and Accurate Reinforcement Learning, as a technique to identify a collection of plausible high-efficacy policies. By drawing potential treatment policies from the variation in current practice---that is, actions currently taken by clinicians---we ensure that our options are likely to be safe, or at least as safe as current practice.

Our results on a cohort of hypotensive ICU patients demonstrate that all three components of SODA-RL (Safety, Diversity, and Quality/Accuracy) are necessary. The distinct policies learned by SODA-RL achieve roughly the same estimated value as the observed clinician policy, and our qualitative results suggest that the different policies do indeed pick up on real underlying options for treatments.

\section*{Background}

We will model the problem of hypotension management as a \emph{Markov Decision Process} (MDP), a standard formalism in reinforcement learning \citep{sutton1998introduction}.  An MDP is defined by a state space $\mathcal{S}$ that describes the current setting of the environment (e.g. clinical variables describing a patient's current physiological state), and an action space $\mathcal{A}$ of possible actions that can be taken (e.g. treatments to administer such as IV fluids or vasopressors).  The Markov in MDP refers to the assumption of Markovianity in the state transition distribiution. That is, we assume that at time $t$, the next state $s_{t+1}$ is determined solely from the current state $s_t$ and action $a_t$, i.e. $p_{\text{trans}}(s_{t+1} | s_{1:t},a_{1:t}) = p_{\text{trans}}(s_{t+1} | s_t, a_t)$, where $s_{1:t}$ and $a_{1:t}$ refer to the complete history of previous states and actions. To complete the specification of the MDP, we define a discount factor $\gamma \in (0,1]$ that balances the value of current vs. future rewards, along with a reward function $r(s,a)$ that assess how good the actions being taken are.  For instance, the reward function might take positive values for physiologically stable states that lead to improved patient outcomes, and take negative values for states leading to physiological instability and decompensation.  

We refer to a decision making strategy as a \textit{policy}, and let $\pi(a|s)$ indicate the probability that action $a$ is taken when in state $s$. In this work we focus on stochastic policies, although it is also possible to learn deterministic policies where the same action is always taken from a given state. A trajectory $\tau$ is a sequence of states, actions and rewards received in an interaction with the environment: $\tau = (s_1, a_1, r_1, \hdots s_{T}, a_{T}, r_{T}$). We define the \textit{value} $V(\pi)$ of a policy as its expected sum of future discounted rewards:
\begin{equation}
V(\pi) = \mathbb{E}_{{\tau \sim p_{\pi,\text{trans}}(\tau)}} \big[ \sum_{t=1}^T \gamma^t r(s_t, a_t) \big],
\label{eqn:value-function}
\end{equation}
where $p_{\pi,\text{trans}}$ denotes the distribution over trajectories $\tau$ generated by following the policy $\pi$ and transitioning between states according to the distribution $p_{\text{trans}}$. 

An optimal policy $\pi^*$ is one that achieves the highest possible value (eq.~\ref{eqn:value-function}). The field of \emph{reinforcement learning} (RL) provides a suite of tools for learning an optimal policy $\pi^*$ via interactions with the environment.  That is, we typically do not have direct access to the transition distribution $p_{\text{trans}}$ and must instead learn by trying actions and seeing their results (e.g. giving a treatment to a patient and observing the outcome).  However, such experimentation is obviously both unethical and impractical in clinical domains, as unsafe actions may be recommended.  The subfield of \emph{batch} RL attempts to learn policies based on previously collected trajectories (e.g. from information in the electronic health record describing the clinical states and treatments given to patients).

A key question in batch RL is \emph{off-policy evaluation}, that is, how to estimate the value of a proposed policy given only a collection of trajectories collected according to some other (potentially suboptimal) policy $\pi_{\text{beh}}$.  One class of methods for accomplishing this relies on importance sampling, a general technique for estimating properties of a distribution of interest (e.g. the distribution of rewards if we follow our policy $\pi$), given only samples generated from a different distribution (e.g. the distribution of rewards if we follow the clinician behavior policy, $\pi_{\text{beh}}$).  In this work, we will use a state-of-the-art estimator, the Consistent Weighted Per-Decision Importance Sampling (CWPDIS, \citep{thomas2015safe}), to estimate the value of the policies we learn using a retrospective set of clinician behavior trajectories $\mathcal{D}_{\text{beh}}$:

\begin{equation}
   V^{\text{CWPDIS}}(
        \pi,
        \mathcal{D}_{\text{beh}}
        )
   \triangleq
   \sum_{t=1}^T \gamma^{t}
    \frac
        {
        \sum_{n \in \mathcal{D}_{\text{beh}}}
            r_{nt}
            \rho_{nt}
        }
        {
        \sum_{n \in \mathcal{D}_{\text{beh}}}
            \rho_{nt}
        }, \; \;  
    \rho_{nt} \triangleq \prod_{i=1}^t
        \frac
        {\pi(a_{ni}|s_{ni})}
        {\pi_{\text{beh}}(a_{ni}|s_{ni})}.
\label{eq:cwpdis}
\end{equation}

The quality of the estimate in eq.~\ref{eq:cwpdis} will depend on how many trajectories are retained by the reweighting by $\rho_{nt}$, known as the \emph{effective sample size} (ESS) \citep{liu1996metropolized}. Informally, the ESS gauges how many samples from the true distribution of interest would provide an estimator with similar quality.  Even though the number of trajectories $N = |\mathcal{D}_{\text{beh}}|$ may be large, high variance in the distribution of the importance weights $\rho_{nt}$ may cause the resulting estimate to be very unreliable, and only provide non-negligible weight on a few trajectories. For instance, if $N=1000$ but the ESS is only $5$, then our estimate using $1000$ trajectories from $\pi_{\text{beh}}$ to estimate the value of our policy $\pi$ will perform about as well as using only $5$ trajectories actually collected according to $\pi$.

We focus on the ESS of the CWPDIS estimator at time $T$, the end of the trajectory:
\begin{equation}
    \text{ESS} = \frac{\left(\sum_{n = 1}^N \rho_{nT} \right)^2}{\sum_{n=1}^N \rho_{nT}}.
\label{eq:ESS}
\end{equation}
If all the importance weights $\{\rho_{nT}\}_{n=1}^N$ are equal, it is easy to see that the ESS is simply $N$. In this work, we use the ESS as an indicator of the reliability of the estimate of a proposed policy's value.  If the ESS is low, then even if the value estimate is high, the proposed policy is not trustworthy and may actually not be high-quality, because that high value  estimate was effectively measured from only a few trajectories. 

\section*{Related Work}

The batch or off-policy RL literature generally focuses on safe and efficient learning using off-policy evaluation techniques \citep{munos2016safe,thomas2015high,gottesman2019guidelines}.  In this work, the notion of safety we use is employed by the assumption that clinicians generally perform well and very rarely make unsafe actions. This is somewhat distinct from other concepts of safety in the area of \textit{safe RL}, such as those comparing the bounds on different off-policy evaluation metrics (e.g. \citep{thomas2015high2}). Moreover, there has been limited exploration of learning collections of distinct agents within the off-policy RL community.

Within RL more broadly, most prior work involves notions of diversity that are not aligned with the kind of efficient exploration-amongst-safe-options setting we are interested in. \citep{liu2017stein,haarnoja2017reinforcement} use notions of diversity that don't directly compare action probabilities, but rather compare features such as neural network parameter differences or the entropy in a single agent's action probabilities. More related is \citep{smith2018inference}, who learn a policy over options and can train multiple options (in an off-policy manner) using a rollout from a single option. Although the distinct options can give rise to agents with distinct behaviors, there is no explicit diversity component in the objective, and it is unclear how to summarize the kinds of distinct trajectories that are possible and what combination of options leads to the most interesting policies. 

Our motivation for seeking a collection of distinct policies in the reinforcement learning setting is aligned most closely with the end goal of \citep{sohrabi2016finding}: presenting a broad set of representative solutions as a tool for hypothesis generation and to discover specific directions of interest for further inquiry.  Their primary application focus is on malware detection, and they first learn a set of good policies followed by a post-hoc clustering step to identify diverse candidates, whereas we learn diverse policies via a joint optimization. \citep{IJCAI-diverseRL} learn collections of distinct policies using a divergence metric between distribution of trajectories induced by policies. However, their work focuses largely on on-policy settings where a simulator of the environment is available and collection of policies is learned sequentially rather than jointly.  In our case, we jointly optimize to find a collection of distinct, plausible alternatives from a collection of already-collected observational data, which can inform clinicians of multiple hypotheses for treatment strategies. 

Finally, there exist several papers using data to inform decisions in the ICU.  \citep{komorowski2018artificial} and \citep{raghu2017continuous} also use RL to learn fluid and vasopressor treatment strategies, but specifically in septic patients, and their focus is on optimality and not safety and diversity.  \citep{girkar2018predicting} focuses on predicting response to fluid bolus therapy, as the treatment does not always work. There are also many papers that attempt to predict onset of  various kinds of interventions (e.g. \citep{ghassemi2017predicting}) and onset of hypotension events (e.g. \citep{hatib2018machine,ghosh2014risk}).  All of these works try to identify one policy, rather than providing reasonable alternatives.

\section*{Cohort and Data Processing}

We draw our trajectories from the publicly-available MIMIC-III database \citep{johnson2016mimic}.  The full database contains static and dynamic information for nearly 60,000 patients treated in the critical care units of Beth-Israel Deaconess Medical Center in Boston between 2001-2012.  We use version 1.4 of MIMIC-III, released in September 2016.  

From the database, we considered adults (at least age 18), with MetaVision data (only patients for whom we could reliably and easily extract both start and end times for interventions). We then removed patients with very short ICU stays of less than 12 hours. For all other ICU stays, we only consider the first 72 hours within the ICU admission, as patients who are in the ICU for extended periods of time often receive different care than the initial treatments in the crucial first few days after admission.  We required at least three distinct measurements of mean arterial pressure (MAP) below 65mmHg, indicating probable hypotension, and used only the first ICU admission if a single patient had multiple admissions.  This filtering process resulted in $10,142$ ICU stays.  We split the dataset into $N=7,000$ ICU stays (of which we use $1,000$ as a validation set for hyperparameter selection and $6,000$ for training), and the remaining $N=3,142$ as a held-out test set for final evaluation.
See Table~\ref{tab:baseline_characteristics} for baseline characteristics and demographics of the selected cohort.

\begin{table}[]
\footnotesize
\begin{tabular}{| l | l |}
\hline \textbf{Characteristic}                              & \textbf{Summary Statistic} \\ \hline \hline
Age, mean years (25/50/75\% quantiles)                              & 67.3 (57.5,69.3,80.5)         \\ \hline
Female (\%)                                             & 47.8\%                   \\ \hline
Surgical ICU (\%)                                       & 48.7\%                   \\ \hline
Non-white (\%)                                          & 23.9\%                   \\ \hline
Emergency Admission                                     & 81.5\%                   \\ \hline
Urgent Admission                                        & 1.2\%                    \\ \hline
Hospital Admit to ICU Admit Time, mean hours (25/50/75\% quantiles)  & 25.7 (0.02, 0.04, 15.97)  \\ \hline 
\end{tabular}
\caption{Baseline characteristics of the total cohort of $N=10,142$ ICU stays we use in this work.}
\label{tab:baseline_characteristics}
\end{table}

In addition to these 7 baseline variables, we also include features derived from 10 different vital signs (e.g. heart rate, MAP) and 20 laboratory measurements (e.g. lactate, creatinine). Vitals are typically recorded about once an hour from (continuous) bedside monitors, while labs are typically only measured a few times a day from blood samples drawn from patients. We also include indicator variables that assess whether or not a variable was recently measured, as the action of decided to measure certain variables may itself be very informative \citep{agniel2018biases}.  

Lastly, we extracted information on the interventions of interest: fluid bolus therapy and vasopressor administrations. We combine different types of fluids and blood products together when forming our fluid action variable (we only include common NaCl 0.9\% solution, lactated ringers, packed red blood cells, fresh frozen plasma, and platelets). We include five different types of vasopressors for the vasopressor action: dopamine, epinephrine, norepinephrine, vasopressin, and phenylephrine. We map these five drugs into a common dosage amount based off norepinephrine equivalents, following the preprocessing in \citep{komorowski2018artificial}, where the infusion rates are in mcg/kg, normalized by body weight.

To apply RL to a problem, we must formalize the state and action spaces, as well as defining a reward and a time-scale.  We now describe each of these pieces below.

\paragraph{State Space, Time Discretization, and Imputation}
We discretize time into hourly windows, and derive an 89-dimensional state vector, consisting of the baseline variables in Table\ref{tab:baseline_characteristics} and values of the physiological and indicator variables as shown in detail in Table\ref{tab:clinical_vars} in the appendix. We impute any unobserved variable with the population median. Once a variable is observed in a given hospital admission, we then use the last observed measurement until a new value is measured. If more than one value is measured in a given hour window we take the most recent value, except for the three blood pressure variables, where we use the minimum value, as clinicians typically treat patients based on their most recent worst blood pressure value.

\paragraph{Action Space} We discretize the two types of interventions, fluid boluses and vasopressors, into 4 and 5 different discrete doses, so that in total there are 20 unique actions (see Figures~\ref{fig:fluid_bins},\ref{fig:vaso_bins},\ref{fig:lowvaso_bins},\ref{fig:overall_actions_bins} in the appendix for details).  To compute the dose of a vasopressor, we aggregate the total amount of vasopressors given in each hour window, normalized by weight. For fluids, we only include fluids boluses of at least 200mL administered in an hour or less. 

\paragraph{Reward} We use the common target of a mean arterial blood pressure (MAP) of 65mmHg.  We consider MAP values above 65mmHg as acceptable (reward 1), and decrease the reward using a piecewise linear function, with inflection points at 60mmHg, and 55mmHg, down to a minimum of 28mmHg (the lowest observed MAP in our data, which we assign a reward of 0). Sufficient urine outputs are allowed to ignore the penalty for moderately low MAP values of 55mmHg or higher, as clinically the slightly lower MAP is less concerning if their fluids are well balanced. See Figure~\ref{fig:reward} in the appendix for a visual depiction of the chosen reward function.  We leave a more thorough investigation of potential reward functions to future work. However, it is important to note that when we present SODA-RL in the next section, rewards are not included in the optimization, so the algorithm will be agnostic to choice of reward and this will only affect the post-hoc value estimates.

\section*{Methods}
When treating hypotension, there may legitimately exist different treatment strategies that are equally effective for a particular patient (e.g. one that focuses on vasopressor use and one that focuses on fluid use). There may also exist treatment strategies whose quality cannot be distinguished from the observational data.

Below, we introduce an algorithm, SODA-RL: Safely Optimized, Diverse, and Accurate Reinforcement Learning, for learning a \emph{collection} of distinct, reliably high-quality policies from a batch of data.  Doing so requires three parts.  First, we want to make sure that any policy ($\pi$) that we recommend \emph{never} takes potentially dangerous actions i.e. $\pi = \text{safe}_{\pi_\text{beh}}(\pi)$.  Second, we want the policy to be high-performing.  Finally, we want the \emph{collection} of policies ($\Pi)$ to be distinct (that is, not repeating the same recommendations).  The following objective function incorporates all of these desiderata:  

\begin{equation}
    \Pi^* = \underset{\Pi}{\text{argmax}} -\mathcal{L}_Q(\pi_{\text{beh}},\Pi) - \lambda \mathcal{L}_D(\Pi), \; \;  \text{s.t.} \; \; \pi = \text{safe}_{\pi_{\text{beh}}}(\pi), \; \; \forall \;  \pi \in \Pi,
\label{eqn:SODA-obj}
\end{equation}
where $\mathcal{L}_Q$ is a loss function that measures discrepancy between our collection of policies and the behavior policy, $\mathcal{L}_D$ is a loss function (with associated regularization strength $\lambda$) related to diversity within the collection of policies. Note that before SODA-RL can be run, we first need to estimate the clinician behavior policy, $\pi_{\text{beh}}$. Following \citep{raghu2018behaviour}, 
we do this using a k nearest neighbors approach to count the proportion of each action observed in the 100 nearest states. To quantify distance between states, we use a manually constructed distance function that weights each of the 89 state variables differently depending on their relative importance to this clinical application.

\paragraph{Safety: $\text{safe}_{\pi_{\text{beh}}}(\pi)$}
The goal of the safety constraint is to ensure that a policy does not take a dangerous action.  For our purpose, we define dangerous as unknown or rarely performed: assuming that the clinicians are choosing amongst reasonable decisions most of the time, there likely exists good reason for treatments that are not chosen.  And even if not, there is no way to tell, given the current data, the potential consequences of a never-tried treatment. 

The safe operator $\text{safe}_{\pi_\text{beh}}$, uses an indicator function ($\mathbbm{1}$) to only allow state-action pairs where the behavior action probability is greater than some threshold $\epsilon$. For a given state, if multiple actions are allowed but some are not, the action probabilities are normalized over only the allowable actions. 
\begin{equation}
    \text{safe}_{\pi_\text{beh}}(\pi(s,a)) \propto \mathbbm{1} [\pi_\text{beh}(s,a) > \epsilon] \cdot \pi(s,a)
\label{eq:safe}
\end{equation}

\paragraph{Distinct ($\mathcal{L}_D(\Pi)$), Likely ($\mathcal{L}_Q(\pi_{\text{beh}},\Pi)$) Collections:}
The safety operation simply ensures that we do not take actions that are completely non-evaluable.  However, it does not ensure that the policies will be of high quality.  One option is to directly optimize policies with respect to the CWPDIS estimator in equation~\ref{eq:cwpdis}.  However, \citep{levine2013guided} note that gradient-based optimization of importance sampling estimates is difficult with complex policies and long rollouts, and we experienced difficulty attempting to optimize this directly.

Thus, we will instead follow a different strategy: our goal will be to identify a collection of \emph{likely, distinct} strategies.  This objective is based on the intuition that the current clinician behaviors are generally reasonable. Our goal is to essentially disentangle the distinct treatment strategies that clinicians are currently using in practice and then each one can be evaluated and filtered using a value estimate from equation ~\ref{eq:cwpdis}.  

We shall measure how likely a proposed policy is given current clinician behavior at a particular state as the difference $l(\pi,\pi_\text{beh})$ where $l$ is some loss function. We will consider the average difference over all policies in the collection $\mathcal{L}^\text{Q}(\pi_{\text{beh}},\Pi)$) and over all states in the batch ($\mathcal{B}$) as the overall similarity (or quality) of the collection of proposed policies and clinician behavior:
\begin{equation}
    L^\text{Q}(\Pi,\pi_{\text{beh}}) = \frac{1}{N} \sum^N_{i=1}  l(\pi_i,\pi_\text{beh}) 
\label{eq:quality}
\end{equation}

Of course, the optimal solution to equation~\ref{eq:quality} is to make all policies in the collection identical to the clinician policy.  To separate out the strategies that clinicians may be using, we add a diversity term, weighted by hyperparameter $\lambda$, that will encourage us to discover a distinct collection of policies.  We define the diversity between two policies as an average of the symmetric KL between their action probabilities, over all observed states in the batch $\mathcal{B}$:
\begin{equation}
    \text{SymKL}^\mathcal{B}_{\pi_i,\pi_j} = \mathbb{E}_{s \sim \mathcal{B}} \left[\frac{1}{2}\text{KL}(\pi_i(s) || \pi_j(s)) + \frac{1}{2}\text{KL}(\pi_j(s) || \pi_i(s))\right]
\label{eq:symKL}
\end{equation}
For a collection of policies $\Pi = \{\pi_1,\pi_2,\hdots,\pi_N\}$, we define the diversity measure as the average of the pairwise diversity measure for pairs that are distinct:
\begin{equation}
    L^\text{D}(\Pi) = \frac{2}{N(N-1)} \sum^N_{i,j : i\neq j} \text{SymKL}^\mathcal{B}_{\pi_i,\pi_j}
\label{eq:diversity}
\end{equation}
Together, equations~\ref{eq:quality} and~\ref{eq:diversity} represent the tension between finding policies that are likely---have high support in the observed data---and yet distinct.  Identifying this collection, we provide a space of potential policies that may be useful in any situation, and the opportunity for clinicians to optimize over the range of action they are already performing.  

\section*{Experimental Setup} 
In this section, we provide details for the setup of our experiments on the particular task of hypotension management in the ICU. We try out two different variants for the loss function $l$ defined in equation~\ref{eq:quality}. The first is the standard cross-entropy (CE) loss function, that will encourage a policy's action probabilities at each state in the batch to be close to the action that was actually taken. The second is the symmetric KL distance (symKL; also used for the diversity term), where here the distance is between the action probabilities for the behavior policy and the policy to be learned. 

In practice, we try a range of $\lambda$ values (1,0.4,0.1,0.01,0.001), and try several values for $\epsilon$ in equation~\ref{eq:safe} (.01,.03,.05; corresponding to only considering actions seen in at least 1, 3, and 5 of the 100 nearest neighbors of a given state, respectively). To actually learn a policy $\pi(a|s)$ that maps states to action probabilities, we use a simple three-layer feedforward neural network (multilayer perceptron), with 128 units per layer. Thus, the parameters to learn are three sets of weight matrices and bias vectors for each policy $\pi$. In our experiments we jointly learn 4 policies at once. We train our methods using the Adam optimizer with a learning rate of $0.001$ and a batch size of 100 trajectories at a time, and use a modest $10^{-6}$ multiplier on an $L_2$ regularization term on all policy parameters.

\paragraph{Evaluation Metrics} 
While our optimization metric aimed to identify distinct, likely treatment policies from the data, our original objective was to identify distinct, effective policies that can serve as options for clinicians.  We evaluate the effectiveness of a policy via the CWPDIS estimator in equation~\ref{eq:cwpdis}, with $\gamma=0.99$.  We also provide the effective sample size of a policy using equation~\ref{eq:ESS}. Together, these metrics provide an estimate of the effectiveness of a policy; CWPDIS value is an estimate of the policy's value, while the ESS is a measure of confidence in that estimate. We also present the CE and symKL loss functions that are optimized in the quality term, as additional metrics to measure how likely a given policy is with respect to the behavior policy and behavior actions taken. We measure the distinctness of the collection using the average symmetric KL between each pair of policies, i.e. equation~\ref{eq:diversity}. Lastly, to measure safety, we count the number of times a policy places a non-negligible probability (i.e. above $0.01$) on an action disallowed by the safety term in equation~\ref{eq:safe}.

\paragraph{Baselines} 

We consider ablations of our approach to determine which aspects are most important to identifying a collection of effective policies. In particular, we explore variants where we turn off various combinations of the diversity and quality terms and safety constraint. We ran experiments with all three (the full method) using both the CE and symKL losses to measure quality, and also ran versions with: only a diversity term with safety constraint, and no quality term; a diversity and quality term but no safety constraint; a quality term and safety constraint, but no diversity term; and diversity term and quality terms alone, with no safety constraints.

\section*{Results}

\begin{table}[]
\scriptsize
\begin{tabular}{|c|l|l|l||l|l|l|l|l|l|l|}
\hline
\multicolumn{1}{|l|}{} & \multicolumn{3}{c||}{\textbf{Setting}} & \multicolumn{7}{c|}{\textbf{Quantitative Metrics}} \\ \hline
\multicolumn{1}{|l|}{} & \begin{tabular}[c]{@{}l@{}}Diversity \\ Weight\end{tabular} & \begin{tabular}[c]{@{}l@{}}Safety \\ Mask?\end{tabular} & \begin{tabular}[c]{@{}l@{}}Quality \\ Term\end{tabular} & \begin{tabular}[c]{@{}l@{}}\# Kept \\ Agents\end{tabular} & \begin{tabular}[c]{@{}l@{}}CWPDIS \\  Value\end{tabular} & \begin{tabular}[c]{@{}l@{}}CE w/ \\ Beh. \\ Actions\end{tabular} & \begin{tabular}[c]{@{}l@{}}SymKL w/ \\ Beh. Action\\ Probabilities\end{tabular} & ESS & \begin{tabular}[c]{@{}l@{}}SymKL btw \\ pairs \\ of agents\end{tabular} & \begin{tabular}[c]{@{}l@{}}\# Times \\ Agents Allowed \\ Unseen Actions\end{tabular} \\ \hline
\multirow{5}{*}{\textit{\begin{tabular}[c]{@{}c@{}}Diverse \\ and Safe\end{tabular}}} & High & Yes & CE & 3 & $34.25\pm0.07$ & $1.03\pm0.04$ & $0.58\pm0.06$ & $352.2\pm94.5$ & $1.95\pm0.21$ & $0\pm0$ \\
 & High & Yes & SymKL & 3 & $35.43\pm1.45$ & $1.13\pm0.11$ & $0.62\pm0.13$ & $221.5\pm102.4$ & $2.05\pm0.23$  & $0\pm0$\\
 & Low & Yes & CE & 0 & - & - & - & - & - & - \\
 & Low & Yes & SymKL & 4 & $36.70\pm0.10$ & $0.52\pm0.00$ & $0.06\pm0.00$ & $282.9\pm30.8$ & $0.00\pm0.00$ & $0\pm0$ \\
 & High & Yes & None & 4 & $35.86\pm1.51$ & $2.44\pm0.65$ & $1.39\pm0.47$ & $310.7\pm180.9$ & $3.27\pm0.00$ & $0\pm0$ \\ \hline
\multirow{5}{*}{\textit{\begin{tabular}[c]{@{}c@{}}Diverse,\\ not Safe\end{tabular}}} & High & No & CE & 0 & - & - & - & - & - & - \\
 & High & No & SymKL & 2 & $41.74\pm0.36$ & $1.14\pm0.15$ & $0.92\pm0.32$ & $234.7\pm146.1$ & $2.90\pm0.00$ & $29230\pm12387$ \\
 & Low & No & CE & 0 & - & - & - & - & - & - \\
 & Low & No & SymKL & 0 & - & - & - & - & - & - \\
 & High & No & None & 0 & - & - & - & - & - & - \\ \hline
\multirow{2}{*}{\textit{\begin{tabular}[c]{@{}c@{}}Safe, not\\ Diverse\end{tabular}}} & None & Yes & CE & 4 & $38.29\pm0.32$ & $0.52\pm0.00$ & $0.08\pm0.00$ & $96.1\pm18.8$ & $0.01\pm0.00$ & $0\pm0$ \\
 & None & Yes & SymKL & 4 & $36.74\pm0.08$ & $0.52\pm0.00$ & $0.06\pm0.00$ & $284.1\pm27.2$ & $0.00\pm0.00$ & $0\pm0$ \\ \hline
\multirow{2}{*}{\textit{\begin{tabular}[c]{@{}c@{}}Not Safe\\ or Diverse\end{tabular}}} & None & No & CE & 0 & - & - & - & - & - & - \\
 & None & No & SymKL & 0 & - & - & - & - & - & - \\ \hline
\end{tabular}
\caption{Quantitative results (means and standard deviations) for each collection of learned policies. For comparison, note that there were $3,142$ trajectories in the test set, so this is the highest achievable ESS. Furthermore, the empirical average of the returns in the test set was $37.90$, so this is a reasonable estimate of the value of the behavior policy. We only show results for agents who learned a policy that had an ESS of at least 50. We show results for $\epsilon=0.03$.}
\label{tab:Results}
\end{table}

Table \ref{tab:Results} presents our quantitative results. As a means of constraining our results to only include policies where we can reliably estimate their value, we prune out learned policies that have an individual ESS of less than 50 on the test set of $N=3142$ trajectories, regardless of their value estimate. In general, most policies that we learn have value estimates that are quite close to the average returns on the test set of $37.90$, which is an unbiased and reliable estimate of the value of the clinician behavior policy.

A major takeaway is that without the safety constraint, the optimization is very likely to end up learning a policy with an unacceptably low ESS. However, even if the ESS is reasonable, there will be a large number of transitions where the agent is recommending unknown, never before seen actions for patients similar to the current state.  Without the diversity but with the safety constraint, it is possible to achieve better CE and SymKL loss values that push you closer to the behavior, but at the cost of very low to no diversity.  Without a quality term of some sort, the combination of diversity and safety learns a very diverse set of policies that still has good value and ESS, but is substantially further away from the behavior.  It often confidently recommends actions that were unlikely, but still possible, under the behavior.  Lastly, using only a quality term also typically fails to learn a policy with a reasonable ESS.  In contrast, the full method SODA-RL using all three terms is a tradeoff in the middle, still learning a fairly diverse set of policies, but sticking much closer to the behavior.

\begin{figure}[t]
\begin{center}
    \includegraphics[width=0.85\linewidth]{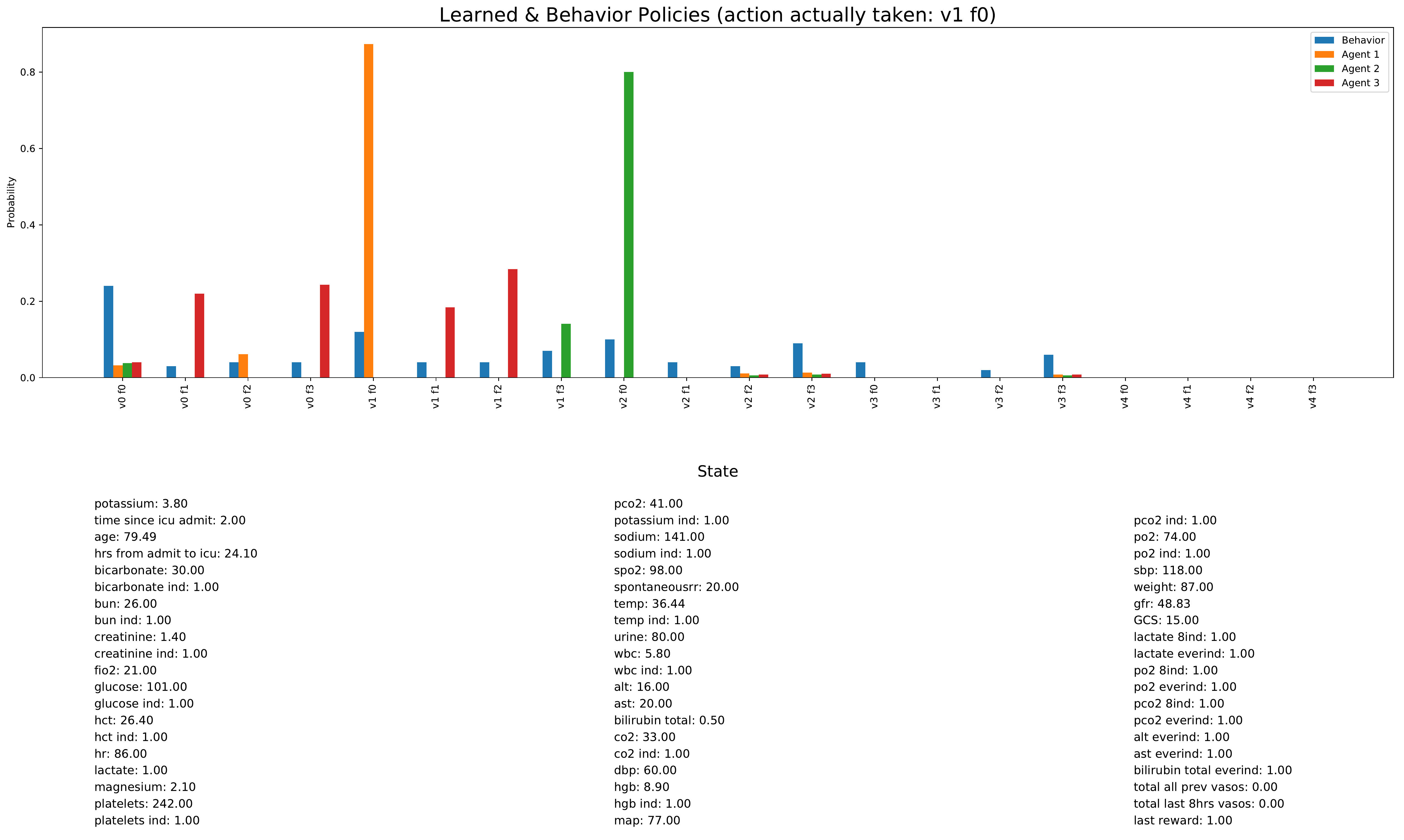}
\end{center}
    \caption{One representative example of a particular state (with variable values presented on the bottom) where the three retained policies learned by SODA-RL exhibit high diversity. For this example, there were 14 actions that the policies were able to exploit, from the safety constraint. Agents 1 and 2 place high probability on a small number of actions, while agent 3 spreads out amongst several reasonable alternatives.}
\label{fig:qual_example}
\end{figure}

\begin{figure}[h]
\begin{center}
    \includegraphics[width=0.85\linewidth]{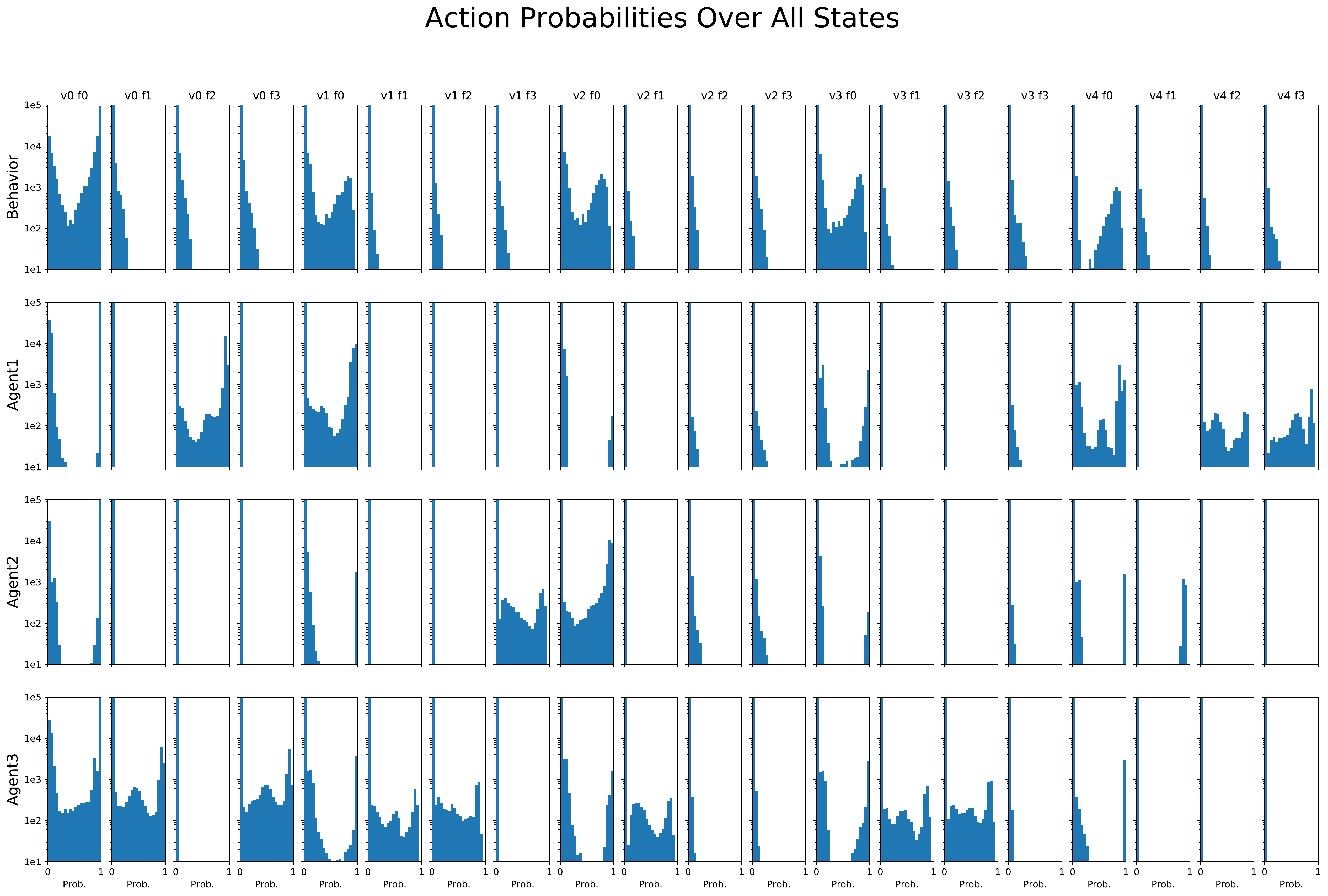}
\end{center}
 \caption{Overall action probabilities. Each column corresponds to one of the 20 actions in our action space. The top row shows the physician behavior probabilities aggregated across all patients in the test set. The bottom three rows show probabilities from the three different agents, from the same run of the algorithm presented in Table~\ref{tab:Results}. }
\label{fig:qual_results_allstates_hist}
\end{figure}

Lastly, we present qualitative results from the policies presented in the second row of Table~\ref{tab:Results}, i.e. high diversity ($\lambda=0.4$), a safety constraint of $\epsilon=0.03$, and the symKL loss in the quality term. Figure~\ref{fig:qual_example} illustrates the local diversity learned by this collection of 3 policies, at a particular state. The blue bars in the figure show the estimated behavior policy action probabilities, while orange, green, and red show the SODA-RL probabilities. Agent 1 (correctly) places high confidence in the low-vasopressor, no-fluid action (v1,f0), while agent 2 places high confidence on the medium-vasopressor, no-fluid action (v2,f0) and agent 3 assigns moderate probability to several other actions.

Figure~\ref{fig:qual_results_allstates_hist} presents a more global picture of the type of diversity that the policies learn. Agent 1 primarily places high probability on high doses of vasopressors with fluids, low doses of vasopressor with no fluids, and medium doses of fluids with no vasopressors. Agent 2 mostly focuses on lower doses of vasopressors, regardless of fluid amount. Lastly, agent 3 largely recommends various amounts of fluids across a range of low to moderate vasopressor doses.

For additional qualitative results similar to these two, see Figures 8-18 in the appendix. Figures 8-14 illustrate additional states with high local diversity at that state among agents, and Figures 15-18 show the distribution of action probabilities across subsets of states where different types of actions were taken and where patients were in states with high physiological instability (i.e. low MAP and high lactate).

\section*{Discussion}

In this paper we introduced SODA-RL, a reinforcement learning approach for identifying a collection of effective treatment policies from observational data.  When applied to the task of hypotension management in the ICU, we found that it is crucial that all three components in Equation~\ref{eqn:SODA-obj} are utilized so that the learned policies are diverse, safe, and not that far from current clinical practice. Additionally, our qualitative results on a learned collection of policies suggests that they are each picking up on diverse sets of practices in the treatment of hypotension. 

 However, one of the major assumptions that we make is that the current set of features that comprise our definition of state are actually sufficient for a clinician to act on (i.e., that our defined state actually satisfies the Markov assumption).  This is likely an unrealistic assumption, but future work could explore other ways of learning state-statistics, and our methods can be seamlessly combined with any state representation.
 
Another interesting line of future work would be to explore how and why different types of vasopressors are given, especially settings where more than one are given (e.g. vasopressin, which is often combined with another drug like norepinephrine). Finally, blood pressure targets themselves are an area of active research \citep{asfar2014high}.  We focused on achieving certain targets in our rewards as that ensures that the actions were closely linked to the outcomes.  More general forms of patient outcomes---e.g. mortality---may be more interesting, but have their own challenges, as these outcomes depend on many factors outside of how a patient's hypotension is managed.  

Overall, we believe SODA-RL represents an important and under-explored direction in reinforcement learning for healthcare: it is often statistically impossible to identify optimal treatment strategies from observational data.  However it \emph{is} possible to identify a collection of plausible alternatives, drawn from current practice variation.  This collection can provide a starting point for clinical experts to perform a targeted review---starting with chart review, perhaps ending in a trial about different treatment options; once vetted, it could be used to help patients and providers think about options in the context of the patient's specific presentation and the provider's experience and expertise.  Our proposed SODA-RL algorithm ensures that those alternatives are distinct and have sufficient support in the data, enabling what we believe will be a more practical and impactful way for clinicians to draw treatment policy insights from observational sources.

\section*{Acknowledgements}
FDV and JF acknowledge support from NSF Project 1750358. MAM and FDV acknowledge support from AFOSR FA 9550-17-1-0155. JF additionally acknowledges Oracle Labs, a Harvard CRCS fellowship, and a Harvard Embedded EthiCS fellowship.

\makeatletter
\renewcommand{\@biblabel}[1]{\hfill #1.}
\makeatother

\bibliographystyle{unsrt}

{\centering \footnotesize
\bibliography{rl_bib}}

\setlength\itemsep{-0.1em}


\newpage
\section*{Appendix}

Our state space contains 89 clinical and demographic features, which we now briefly describe. There are 7 baseline variables (demographics and other characteristics available on ICU admission) in Table~\ref{tab:baseline_characteristics}. We also include in the state formulation a continuous variable denoting how far into the first 72 hours of ICU stay a current time point is. The remaining 81 clinical variables are in Table~\ref{tab:clinical_vars}, which summarizes the measured value of time series variables as well as the indicator variables. Lastly, there are indicator variables for the most recent type of treatment administered, and for the total amount of each treatment administered thus far and in the last 8 hours.

\begin{table}[]
\tiny
\begin{tabular}{| l | l |}
\hline
\textbf{Clinical Variable} & \begin{tabular}[c]{@{}l@{}} \textbf{Summary Statistic:} \\ \textbf{Mean (25/50/75\% quantiles),} \\ \textbf{or Percentage of Times} \end{tabular}  \\ \hline 
 Bicarbonate                                          &   24.3 (21.0, 24.0, 27.0)               \\
 Bicarbonate, indicator if measured last hour            &    14.2\%                \\
 BUN                                          &      27.9 (14.0, 21.0, 34.0)             \\
 BUN, indicator if measured last hour            &    8.5\%                \\
 Creatinine                                          &      1.5 (0.7, 1.0, 1.6)             \\
 Creatinine, indicator if measured last hour            &   8.5\%                 \\
 GFR                                          &  69.9 (36.5, 63.7, 94.8)               \\
 FiO2                                          &   54.9 (40.0, 50.0, 60.0)               \\
 FiO2, indicator if measured last hour            &    13.4\%               \\
 FiO2, indicator if ever measured            &  61.6\%                 \\
 Glucose                                          &     139.6 (106.0, 127.0, 156.0)               \\
 Glucose, indicator if measured last hour            &   28.6\%              \\
 Hct                                          &     30.6 (26.9, 30.0, 33.8)             \\
 Hct, indicator if measured last hour            &  10.8\%                 \\
 HR                                          &    84.6 (72.0, 83.0, 96.0)             \\
 HR, indicator if  measured last hour            &   94.4\%                 \\
 Lactate                                          &   2.6 (1.3, 1.9, 3.0)                 \\
 Lactate, indicator if measured last hour            &   5.2\%                 \\
 Lactate, indicator if measured in last 8 hours            &    28.3\%                \\
 Lactate, indicator if ever measured         &    78.1\%               \\
 Magnesium                                          &      2.1 (1.8, 2.0, 2.3)              \\
 Magnesium, indicator if  measured last hour            &   6.9\%                 \\
 Platelets                                          &  200.0 (125.0, 182.0, 250.0)                \\
 Platelets, indicator if  measured last hour            & 8.1\%                  \\
 Potassium                                          &  4.2 (3.8, 4.1, 4.5)                 \\
 Potassium, indicator if  measured last hour            &  11.6\%                 \\
 Sodium                                          &   138.1 (135.0, 138.0, 141.0)                 \\
 Sodium, indicator if  measured last hour            &    9.8\%                \\
 SPO2                                          &  96.8 (95.0, 97.0, 99.0)                \\
 SPO2, indicator if  measured last hour            &  92.2\%                \\
 Spontaneous RR                                          &   19.4 (16.0, 19.0, 23.0)                 \\
 Spontaneous RR, indicator if  measured last hour            &   93.9\%               \\
 Temp                                          &      36.9 (36.4, 36.8, 37.4)             \\
 Temp, indicator if  measured last hour            &     28.9\%               \\
  Urine Output in last hour                                         &  115.0 (40.0, 75.0, 140.0)                 \\
 Urine Output, indicator if  measured last hour            &   63.5\%               \\
 WBC                                          &    11.9 (7.6, 10.5, 14.4)                \\
 WBC, indicator if  measured last hour            &   7.9\%               \\
 ALT                                          &    212.3 (18.0, 32.0, 79.0)               \\
 ALT, indicator if  measured last hour            &   2.6\%             \\
 ALT, indicator if ever measured           &    66.5\%                \\
 AST                                          &    285.7 (25.0, 44.0, 112.0)               \\
 AST, indicator if  measured last hour            & 2.6\%                  \\
 AST, indicator if ever measured           & 66.5\%                 \\
 Bilirubin Total                                          &   1.375 (0.5, 0.9, 0.9)                \\
 Bilirubin Total, indicator if  measured last hour            &  3.6\%                 \\
 Bilirubin Total, indicator if ever measured           &   66.1\%                 \\
 CO2                                          &   24.5 (22.0, 24.0, 27.0)                 \\
 CO2, indicator if  measured last hour            &    8.3\%                \\
 DBP                                          &  57.2 (49.0, 56.0, 64.0)                 \\
 DBP, indicator if  measured last hour            &   91.5\%                \\
 Hgb                                          &    10.4 (9.1, 10.2, 11.6)                \\
 Hgb, indicator if  measured last hour            &    13.1\%                \\
 MAP                                          &   72.8 (64.0, 71.0, 80.0)                 \\
 MAP, indicator if  measured last hour            &  91.9\%                  \\
 PCO2                                          &   41.4 (35.0, 40.0, 46.0)                \\
 PCO2, indicator if  measured last hour            & 8.3\%                   \\
 PCO2, indicator if measured in last 8 hours            &    35.3\%              \\
 PCO2, indicator if ever measured           &  70.2\%                  \\
 PO2                                          &      149.9 (86.0, 117.0, 176.0)              \\
 PO2, indicator if  measured last hour            &  8.3\%                  \\
 PO2, indicator if measured in last 8 hours            &  35.3\%                 \\
 PO2, indicator if ever measured           &     70.2\%              \\
 SBP                                          &  113.6 (100.0, 112.0, 126.0)                 \\
 SBP, indicator if  measured last hour            &   91.6\%                 \\
 Weight                                          &  83.6 (66.8, 80.0, 96.7)                \\
 Weight, indicator if  measured last hour            &   3.6\%                \\
 GCS                                          &    12.0 (10.0, 14.0, 15.0)               \\
 GCS, indicator if  measured last hour            &   28.3\%                \\
 Indicator for if vasopressor action 1 administered last hour   &  5.1\%                  \\
 Indicator for if vasopressor action 2 administered last hour   &  5.8\%                 \\
 Indicator for if vasopressor action 3 administered last hour   &  4.7\%                 \\
 Indicator for if vasopressor action 4 administered last hour   &  2.4\%                 \\
 Indicator for if fluid action 1 administered last hour   &  1.7\%                  \\
 Indicator for if fluid action 2 administered last hour   &   2.1\%                \\
 Indicator for if fluid action 3 administered last hour   &  1.9\%                  \\
 Total amount of vasopressor administered during ICU stay   &   129.7 (0.0, 0.0, 51.7)                \\
 Total amount of fluids administered during ICU stay   &   1891.0 (0.0, 850.0, 2940.7)                \\
 Total amount of vasopressor administered in last 8 hours   &  28.6 (0.0, 0.0, 0.0)                 \\
 Total amount of fluids administered in last 8 hours   &  342.1 (0.0, 0.0, 202.18)               \\
 Reward value during last hour   &   0.978 (1.000, 1.000, 1.000) \\ \hline
\end{tabular}            
 \caption{Summary statistics for clinical variables included in our state formulation. Continuous-valued time series are summarized by their mean, and 25/50/75\% quantiles, among measured values (i.e. excluding imputed values). Indicators are summarized by the percentage of hours in the full dataset where they were active.}
 \label{tab:clinical_vars}
\end{table}


\clearpage
\newpage

We now present histograms showing the distribution of actual values of treatments given, to show how we eventually discretized them to achieve our final action space of 20 possible actions. 

\begin{figure}[t]
\begin{center}
    \includegraphics[width=0.5\linewidth]{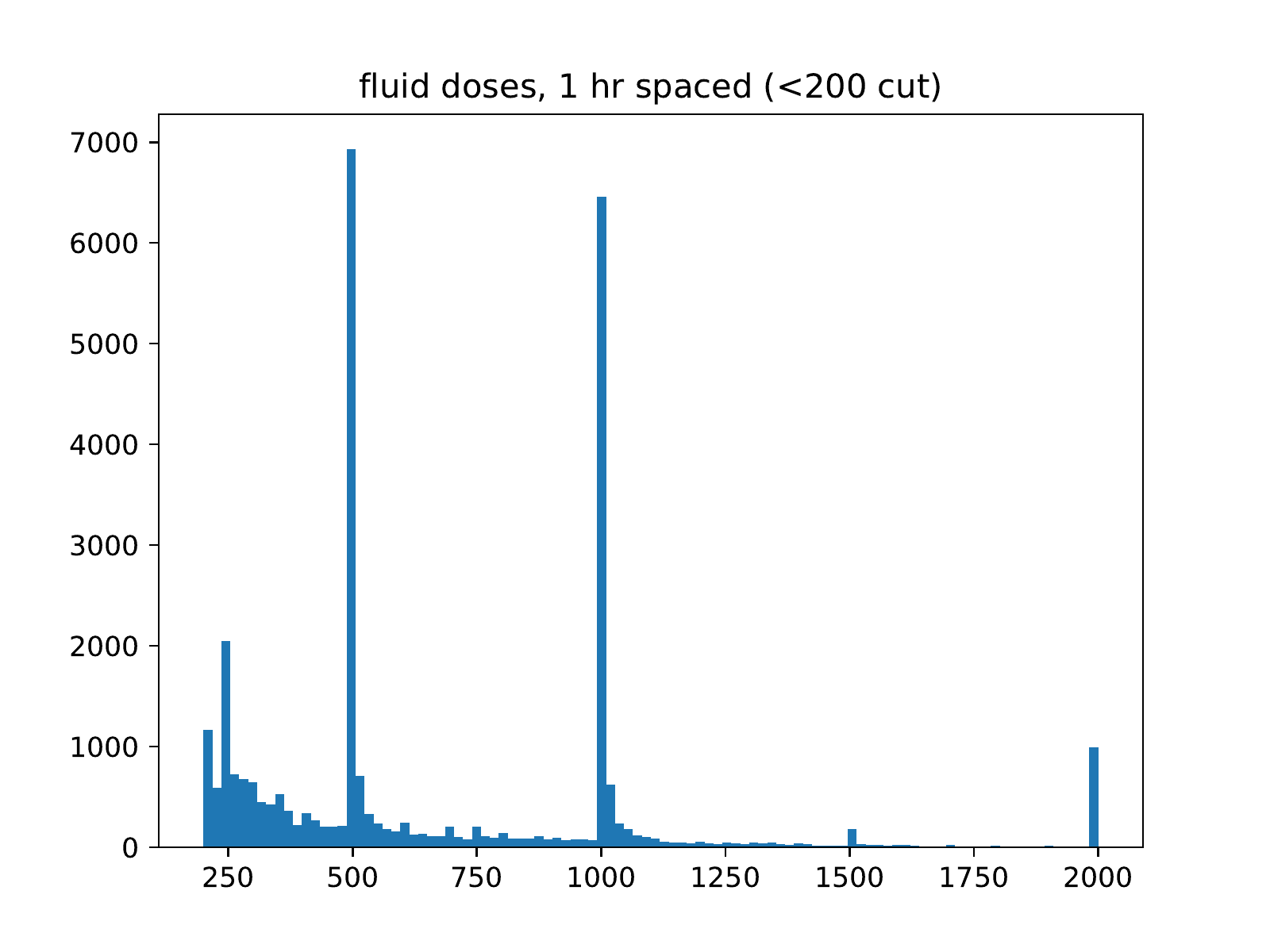}
\end{center}
    \caption{Histogram showing the raw observed values of fluid boluses administered in the dataset. We define a fluid bolus to be IV fluids administered within an hour, and a volume of at least 200mL. From this plot, we decided to discretize fluids to the following 4 discrete ranges (in mL within an hour): $\{[0,200), [200,500), [500,1000), [1000,2000)\}$.}
\label{fig:fluid_bins}
\end{figure}

\begin{figure}[t]
\begin{center}
    \includegraphics[width=0.5\linewidth]{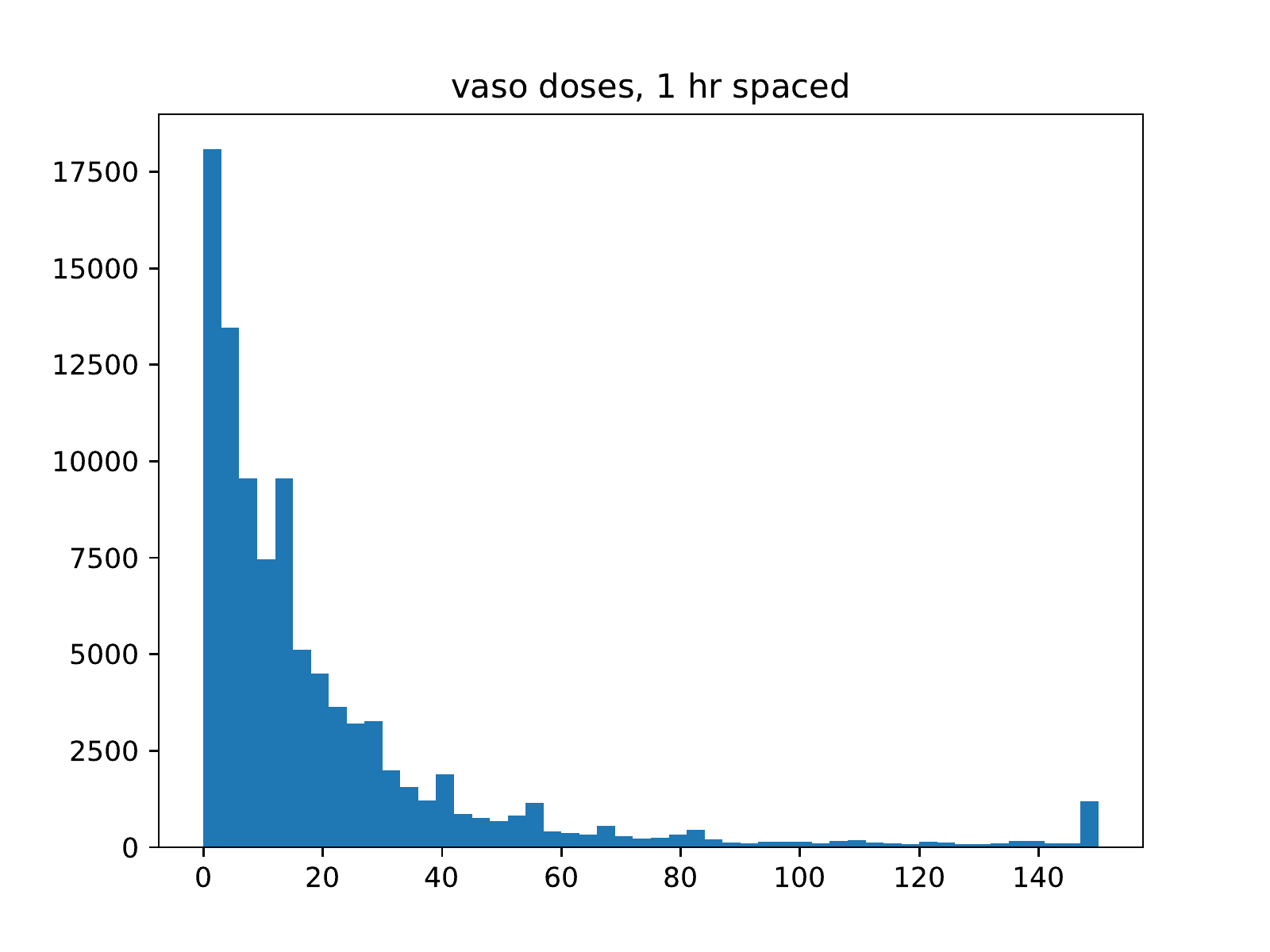}
\end{center}
    \caption{Histogram showing the raw observed amounts of vasopressor administered within each discrete hour in the dataset.  From this plot, together with the next plot that zooms in on values less than 40, we decided to discretize vasopressors to the following 5 discrete ranges: $\{0, (0,5), [5,15), [15,40), [40,150)\}$.}
\label{fig:vaso_bins}
\end{figure}

\begin{figure}[t]
\begin{center}
    \includegraphics[width=0.5\linewidth]{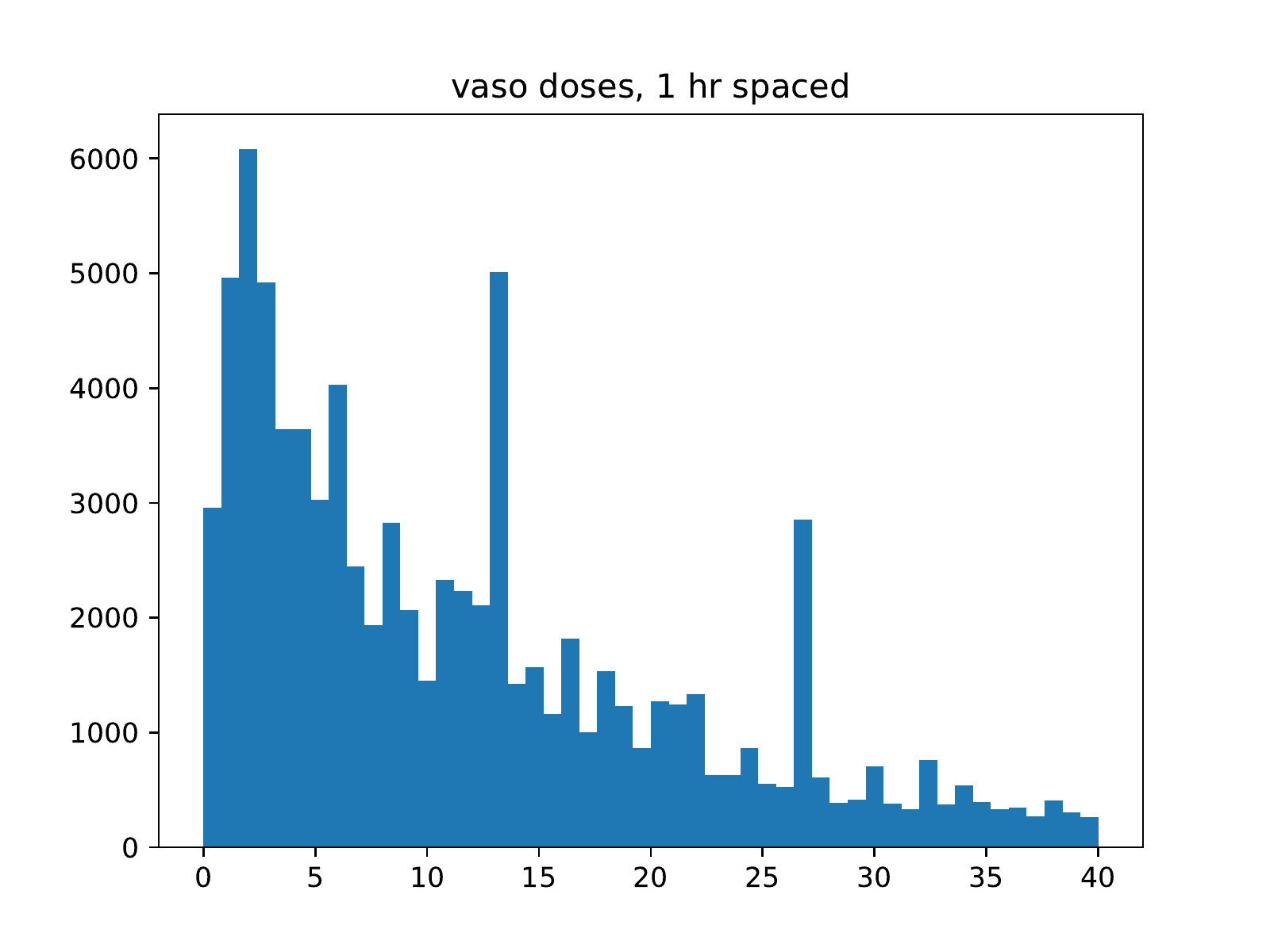}
\end{center}
    \caption{Histogram showing the lower range of the raw observed amounts of vasopressor administered within each discrete hour in the dataset.  This is just a zoomed in version of Figure \ref{fig:vaso_bins}.}
\label{fig:lowvaso_bins}
\end{figure}

\begin{figure}[t]
\begin{center}
    \includegraphics[width=0.5\linewidth]{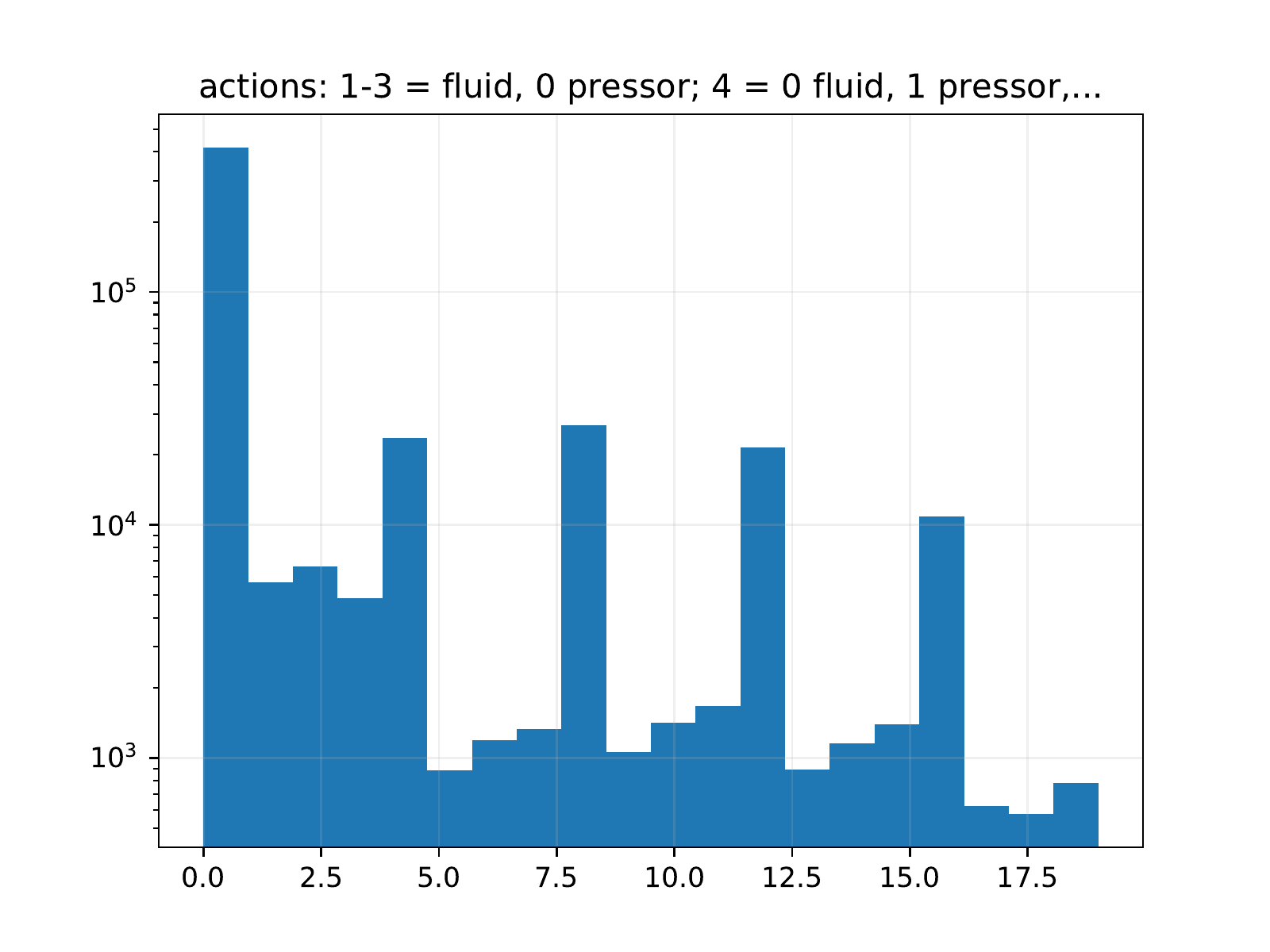}
\end{center}
    \caption{Histogram of overall counts of the 20 different actions in the entire dataset. Actions 0,1,2,3 are no vasopressor and none/low/medium/high fluid bolus; 4,5,6,7 are low dose of vasopressor along with no/low/medium/high fluid bolus, etc.}
\label{fig:overall_actions_bins}
\end{figure}

\begin{figure}[t]
\begin{center}
    \includegraphics[width=0.5\linewidth]{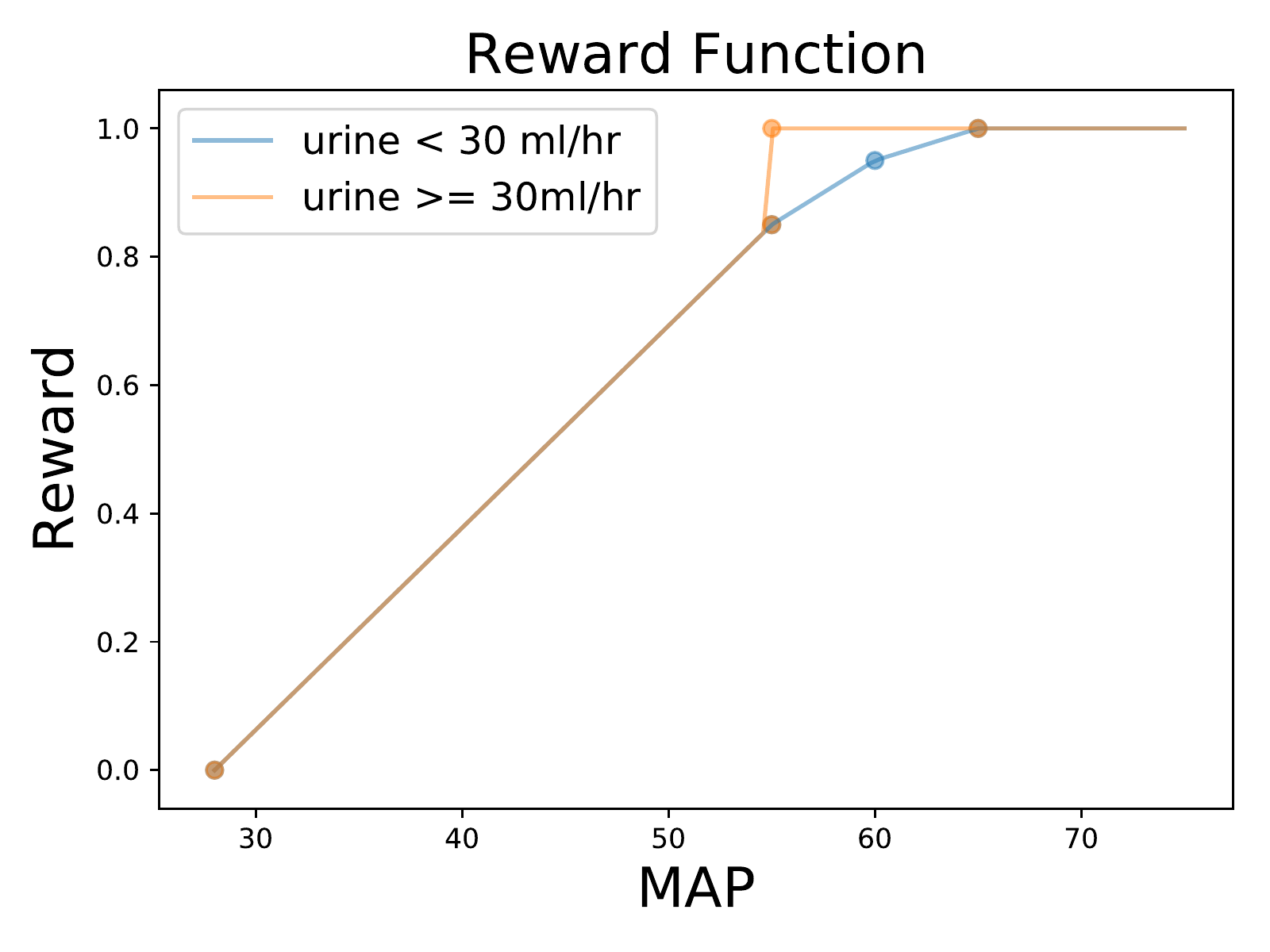}
\end{center}
    \caption{Reward function used in our analysis. The reward depends primarily on MAP, with inflection points at 55, 60, and 65. Values 65 and above are considered optimal. Note that the main objective function for SODA-RL does not depend on the specific reward function chosen, so this will only affect our final estimate of the overall value of the learned policies.}
\label{fig:reward}
\end{figure}

\clearpage
\newpage
We now show additional results figures exploring different specific states observed in the test set, where the three retained policies learned by SODA-RL exhibit high degrees of diversity.

\begin{figure}[t]
\begin{center}
    \includegraphics[width=0.8\linewidth]{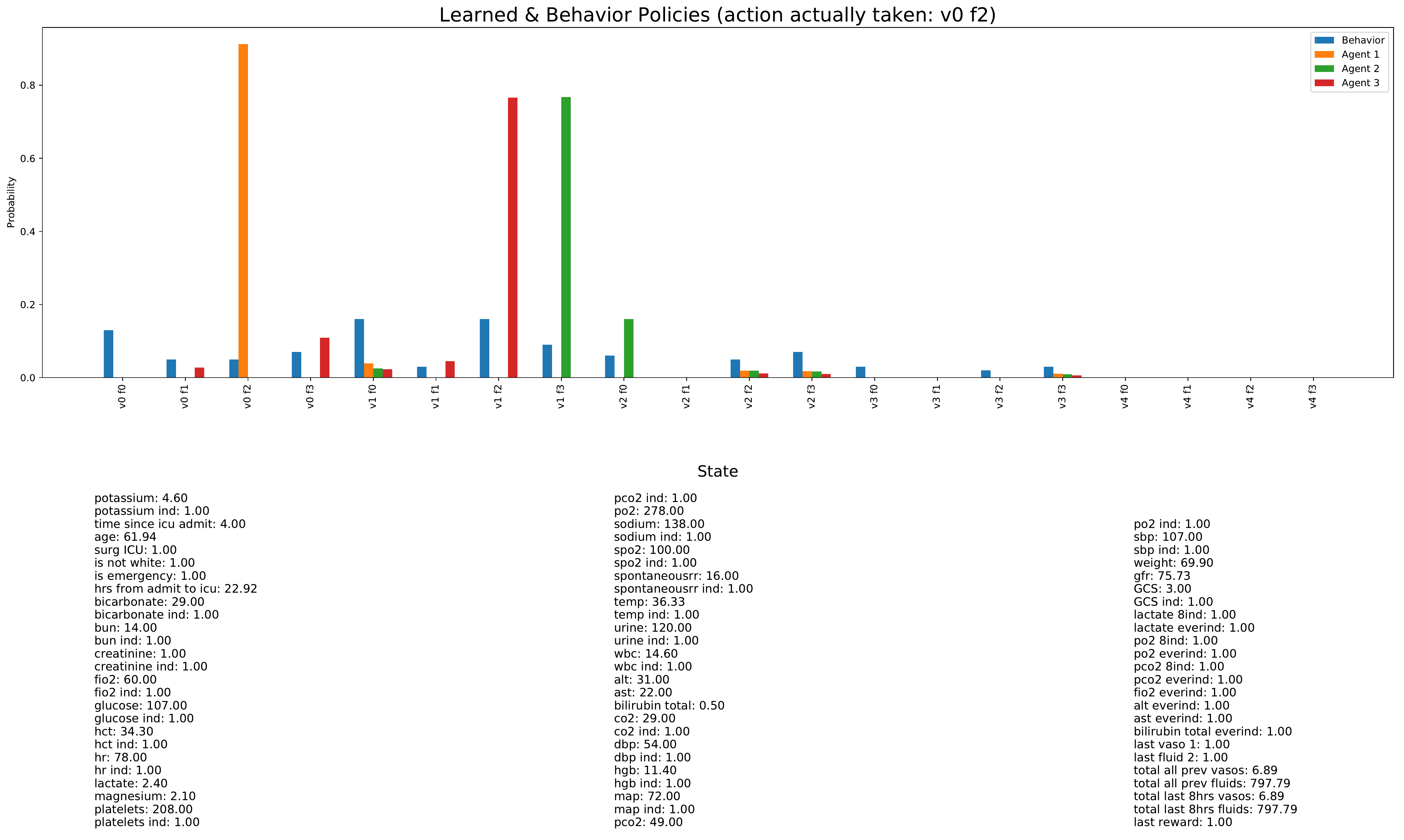}
\end{center}
    \caption{An example state where there were 13 different actions allowed by the safety term in SODA-RL, and where the resulting policies exhibit high diversity.}
\label{fig:}
\end{figure}

\begin{figure}[t]
\begin{center}
    \includegraphics[width=0.8\linewidth]{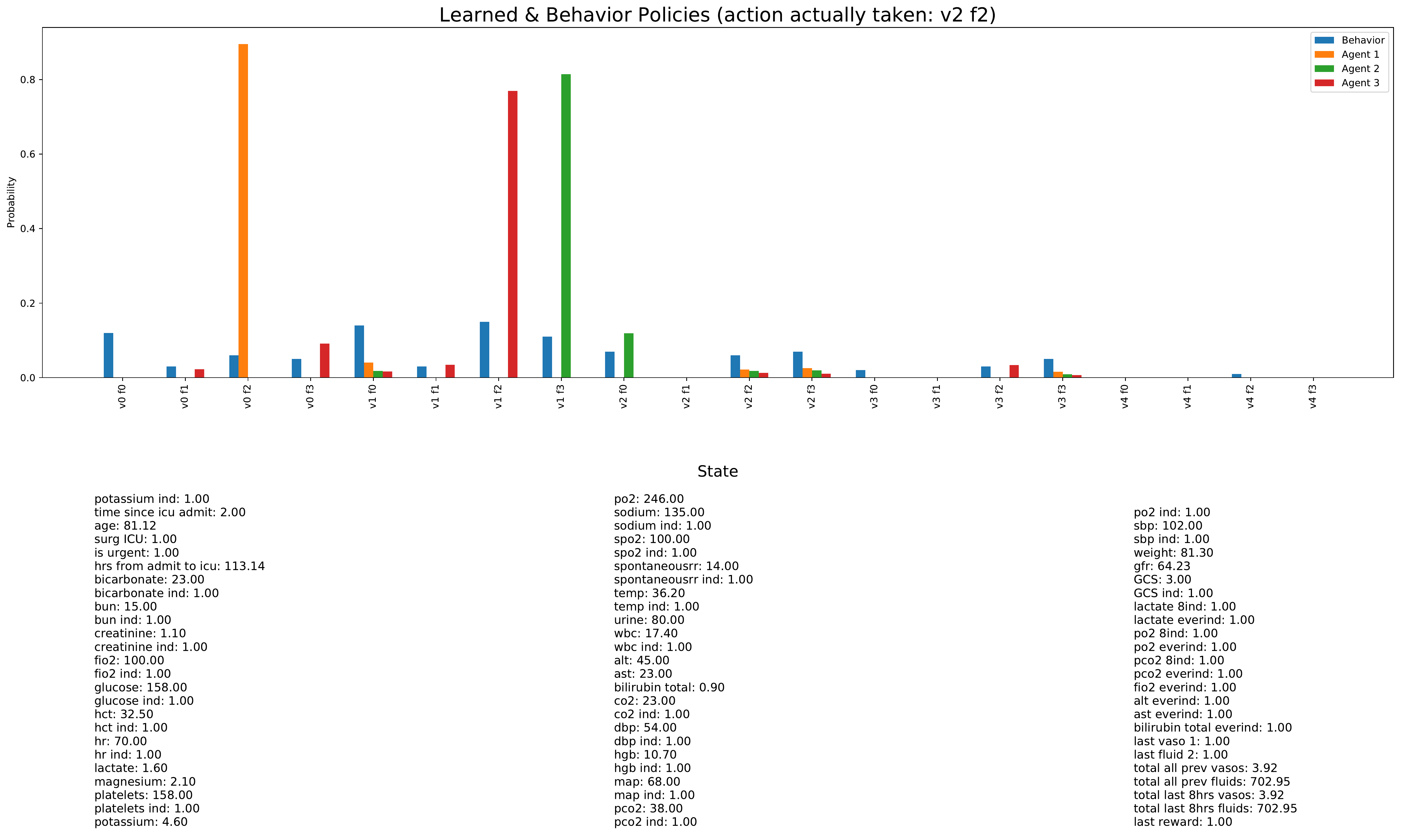}
\end{center}
    \caption{An example state where there were 13 different actions allowed by the safety term in SODA-RL, and where the resulting policies exhibit high diversity.}
\label{fig:}
\end{figure}

 \begin{figure}[t]
 \begin{center}
     \includegraphics[width=0.8\linewidth]{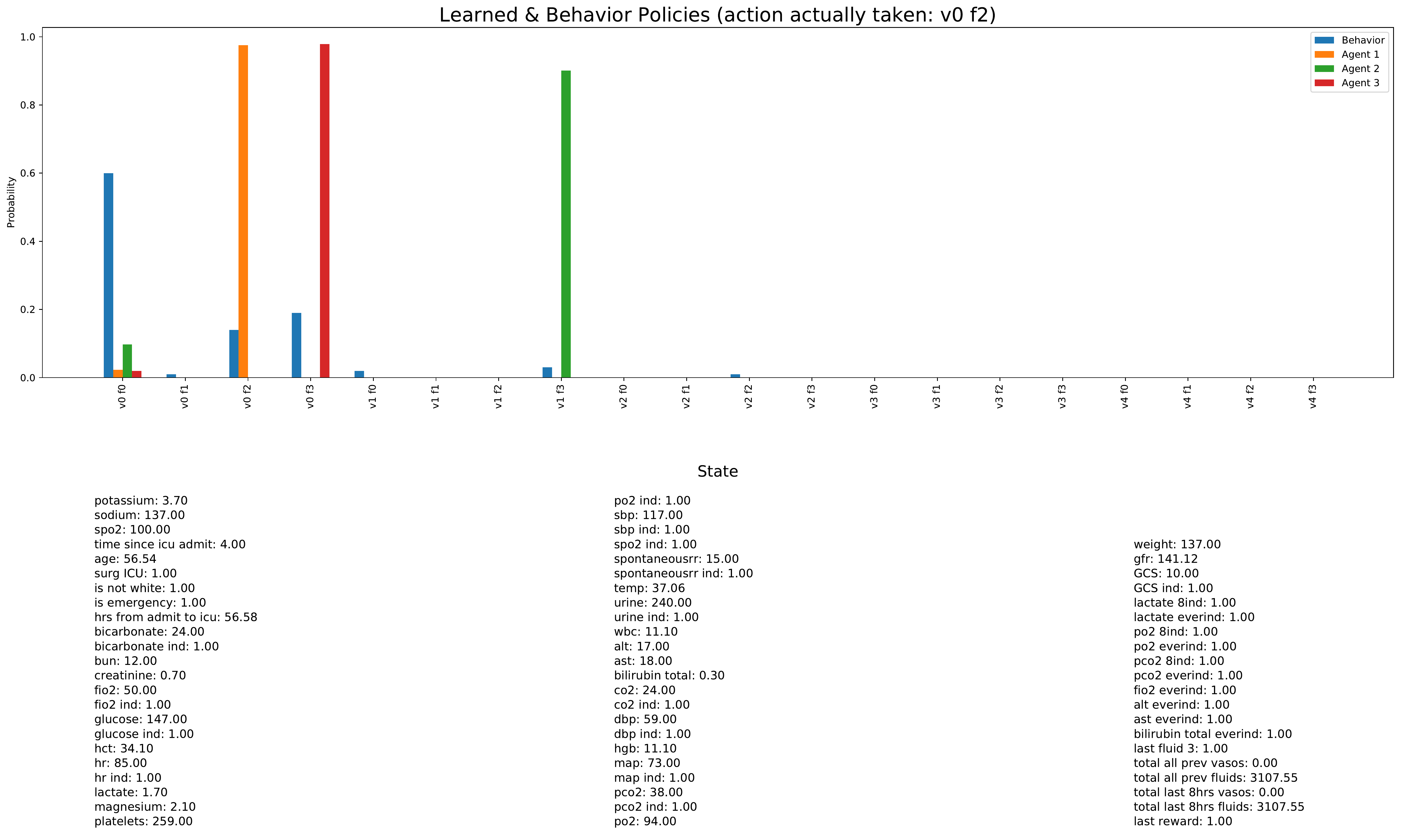}
 \end{center}
     \caption{An example state where there were 4 different actions allowed by the safety term in SODA-RL, and where the resulting policies exhibit high diversity.}
 \label{fig:}
 \end{figure}

 \begin{figure}[t]
 \begin{center}
     \includegraphics[width=0.8\linewidth]{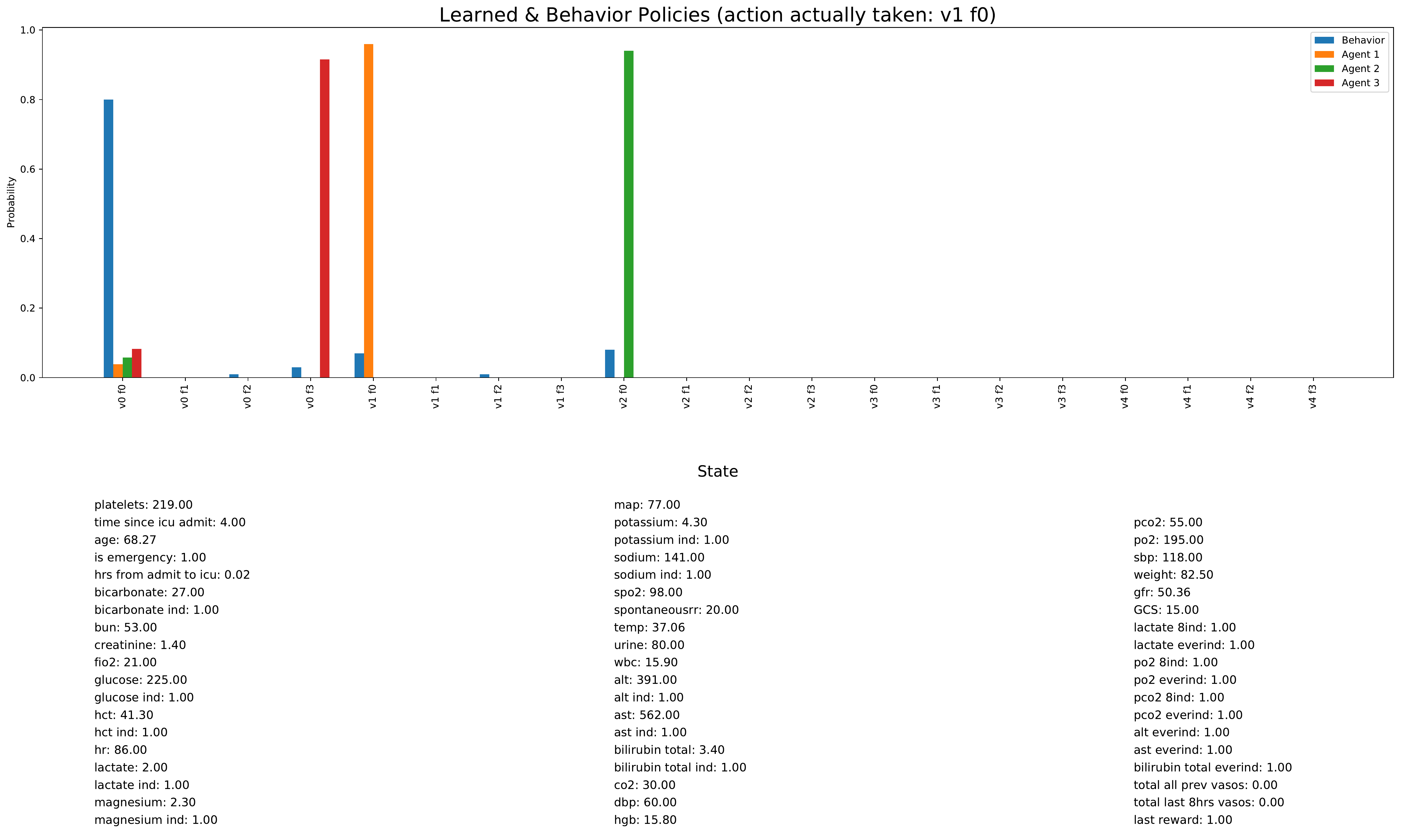}
 \end{center}
     \caption{An example state where there were 4 different actions allowed by the safety term in SODA-RL, and where the resulting policies exhibit high diversity.}
 \label{fig:}
 \end{figure}

 \begin{figure}[t]
 \begin{center}
     \includegraphics[width=0.8\linewidth]{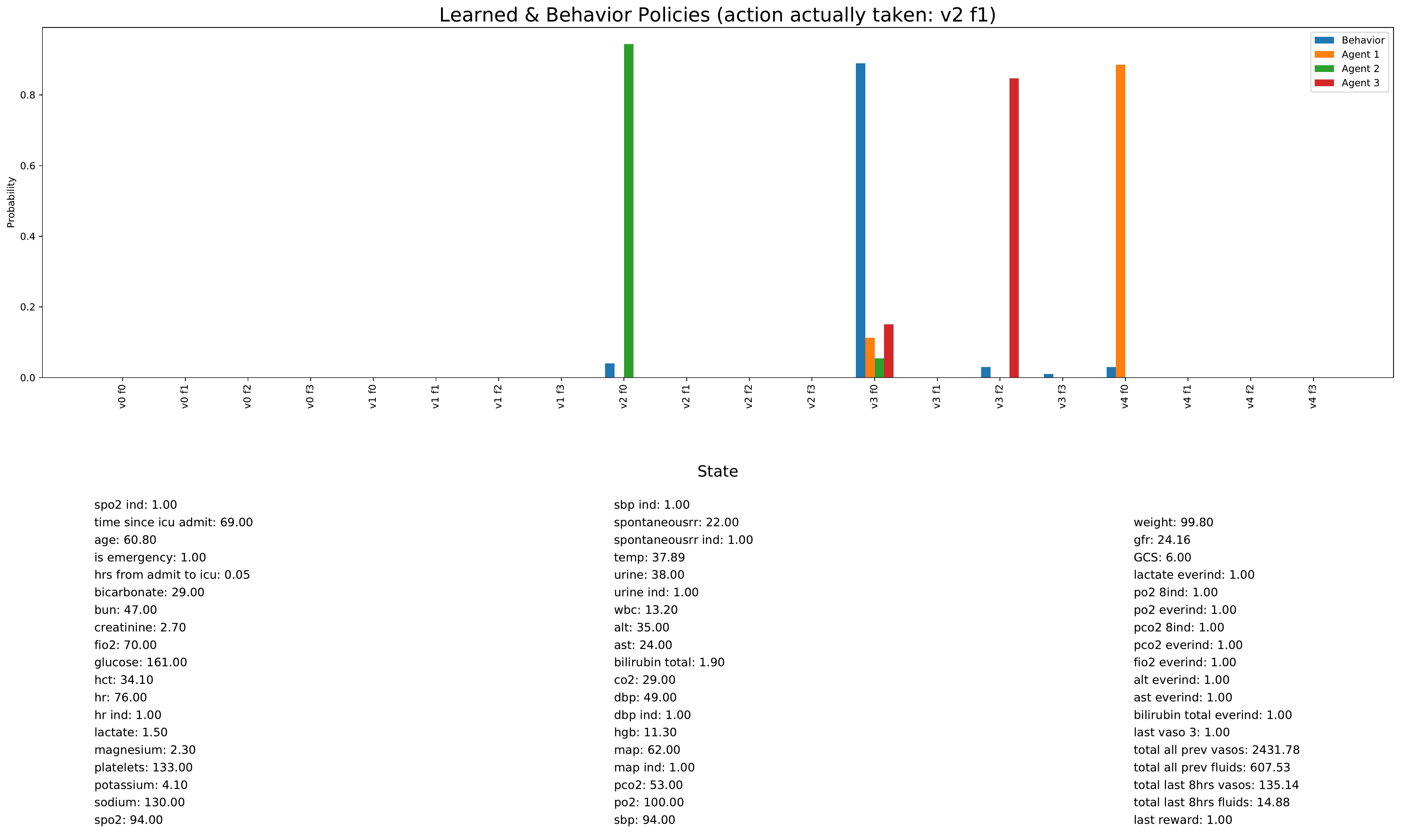}
 \end{center}
     \caption{An example state where there were 4 different actions allowed by the safety term in SODA-RL, and where the resulting policies exhibit high diversity.}
 \label{fig:}
 \end{figure}

 \begin{figure}[t]
 \begin{center}
     \includegraphics[width=0.8\linewidth]{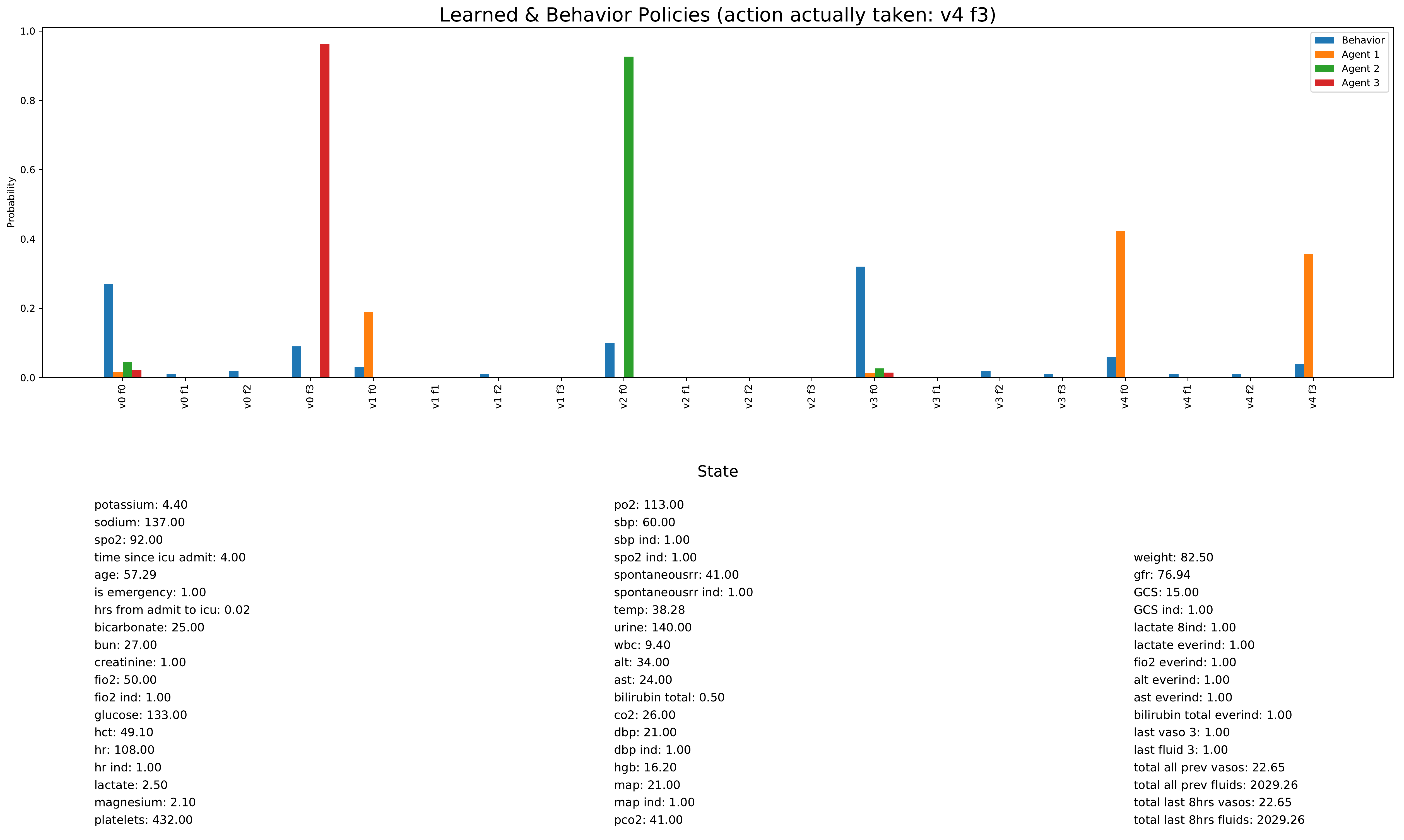}
 \end{center}
     \caption{An example state where there were 7 different actions allowed by the safety term in SODA-RL, and where the resulting policies exhibit high diversity.}
 \label{fig:}
 \end{figure}

 \begin{figure}[t]
 \begin{center}
     \includegraphics[width=0.8\linewidth]{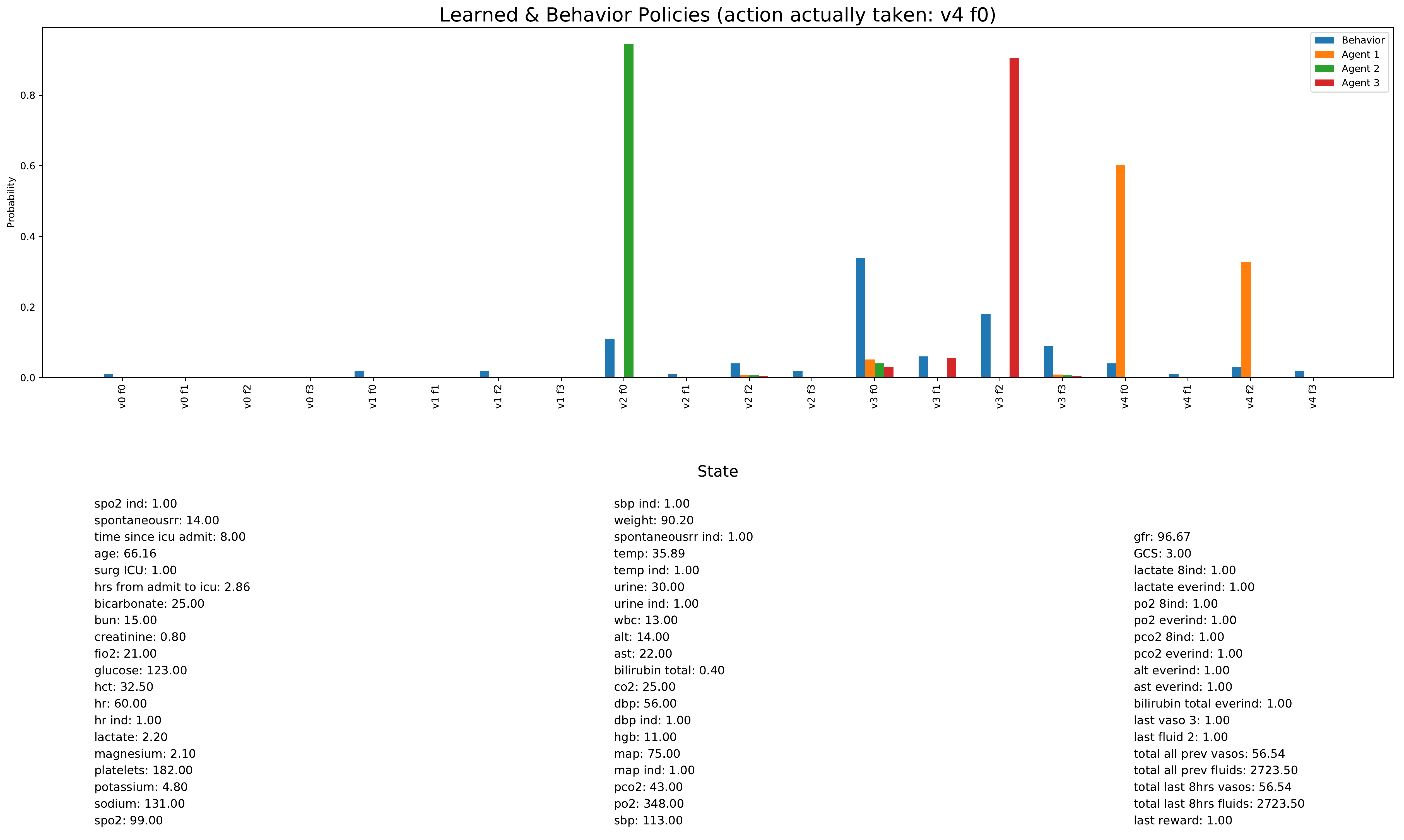}
 \end{center}
     \caption{An example state where there were 8 different actions allowed by the safety term in SODA-RL, and where the resulting policies exhibit high diversity.}
 \label{fig:}
 \end{figure}

\begin{figure}[t]
 \begin{center}
     \includegraphics[width=1\linewidth]{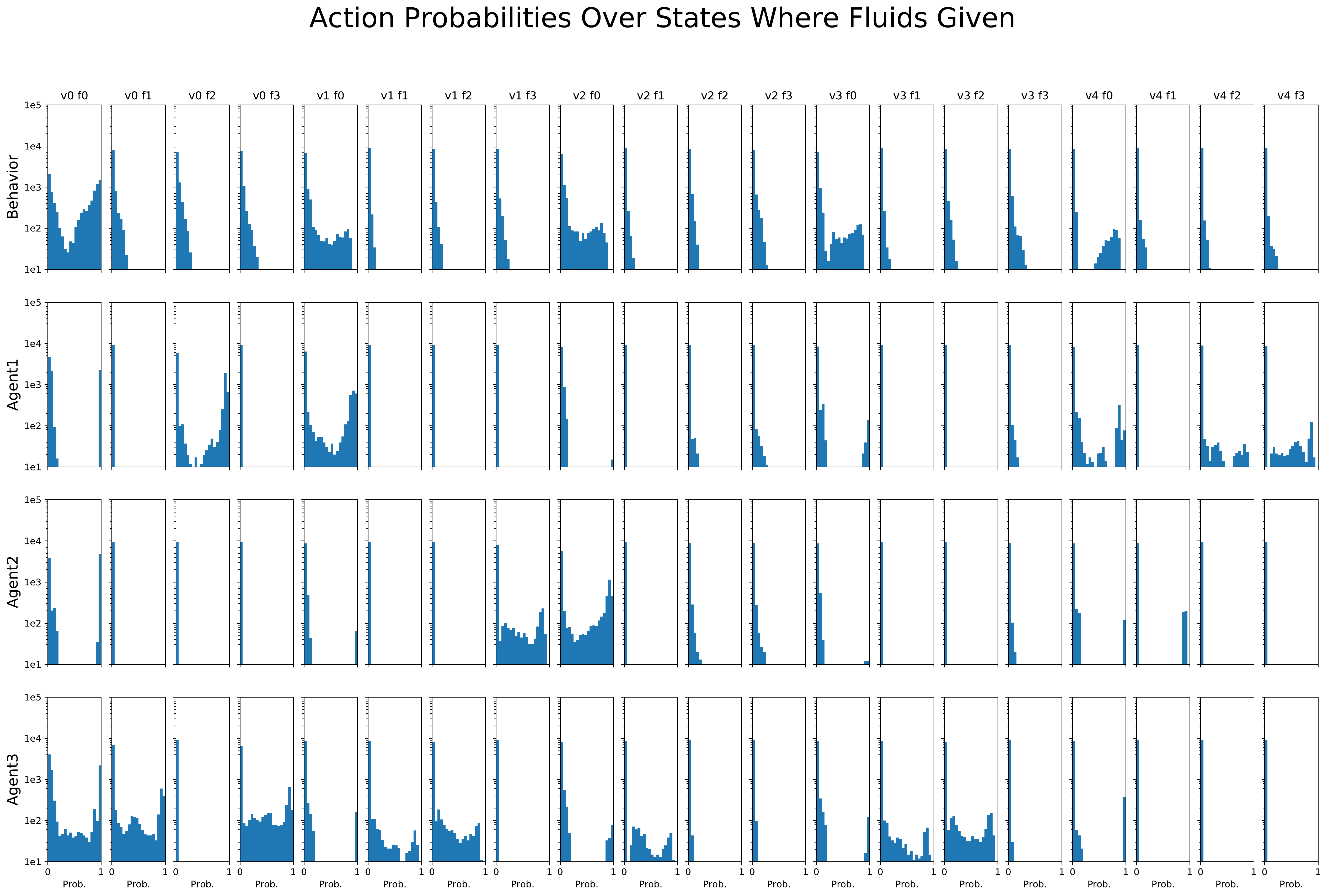}
 \end{center}
  \caption{Action probabilities for states where fluids are subsequently administered. Each column corresponds to one of the 20 actions in our action space. The top row shows the physician behavior probabilities aggregated across all patients in the test set. The bottom three rows show probabilities from the three different agents, from the same run of the algorithm presented in the results table \ref{tab:Results}.}
 \label{fig:fluid_actions_hist}
 \end{figure}
\begin{figure}[t]
 \begin{center}
     \includegraphics[width=1\linewidth]{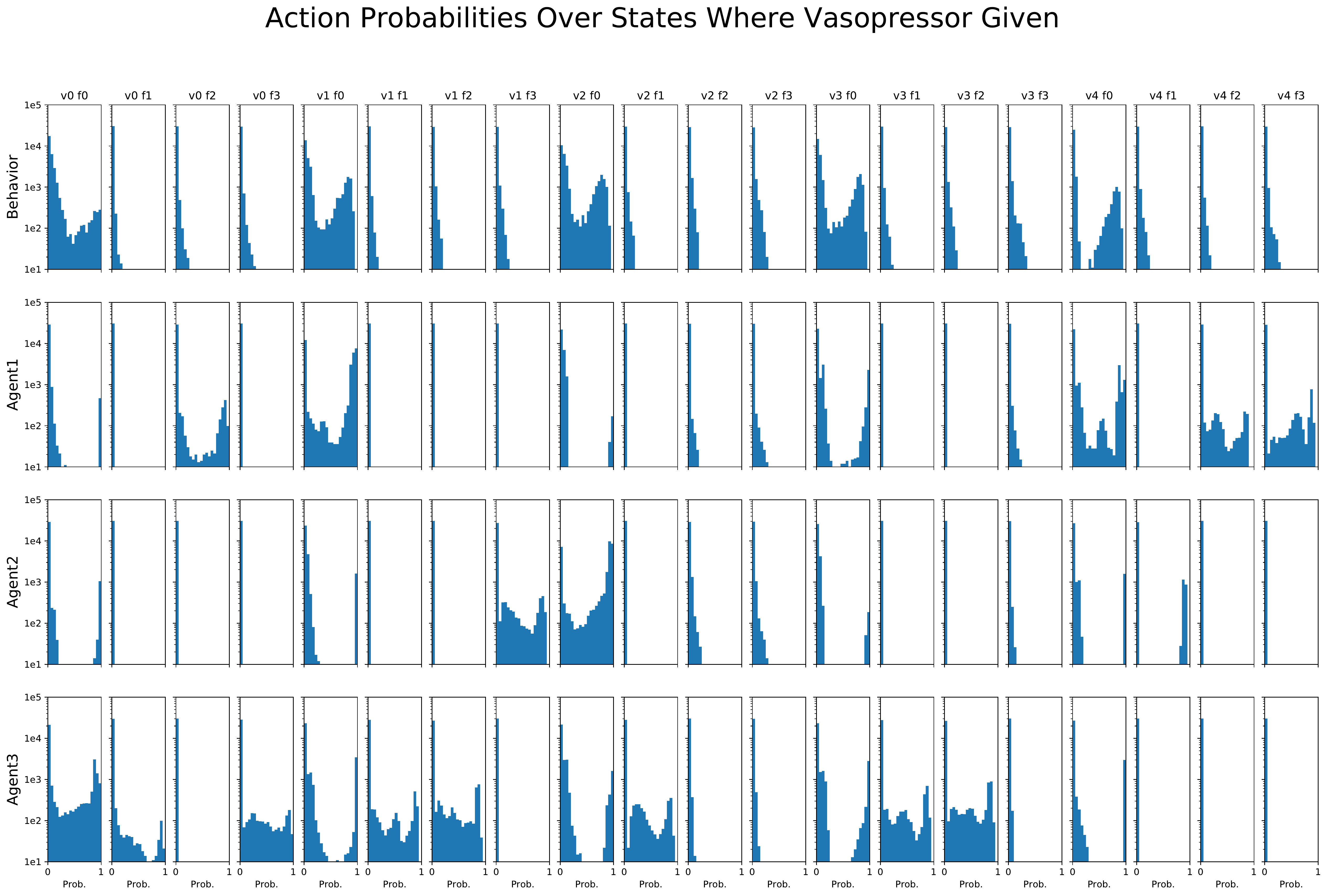}
 \end{center}
  \caption{Action probabilities for states where a vasopressor is subsequently administered. Each column corresponds to one of the 20 actions in our action space. The top row shows the physician behavior probabilities aggregated across all patients in the test set. The bottom three rows show probabilities from the three different agents, from the same run of the algorithm presented in the results table \ref{tab:Results}.}
 \label{fig:vaso_actions_hist}
 \end{figure}
\begin{figure}[t]
 \begin{center}
     \includegraphics[width=1\linewidth]{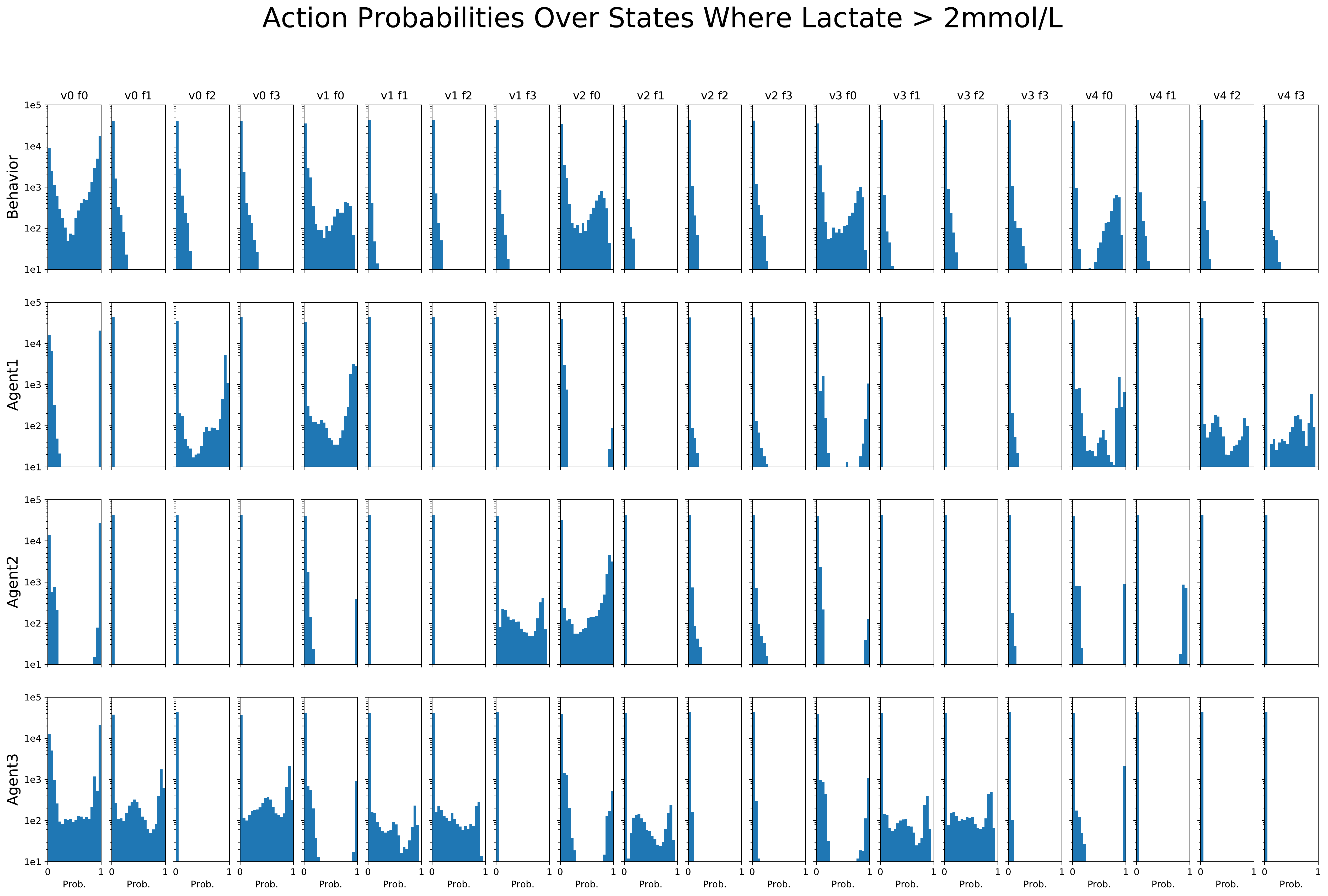}
 \end{center}
  \caption{Action probabilities for states with an elevated lactate of greater than 2mmol/L. Each column corresponds to one of the 20 actions in our action space. The top row shows the physician behavior probabilities aggregated across all patients in the test set. The bottom three rows show probabilities from the three different agents, from the same run of the algorithm presented in the results table \ref{tab:Results}.}
 \label{fig:highlactate_actions_hist}
 \end{figure}
\begin{figure}[t]
 \begin{center}
     \includegraphics[width=1\linewidth]{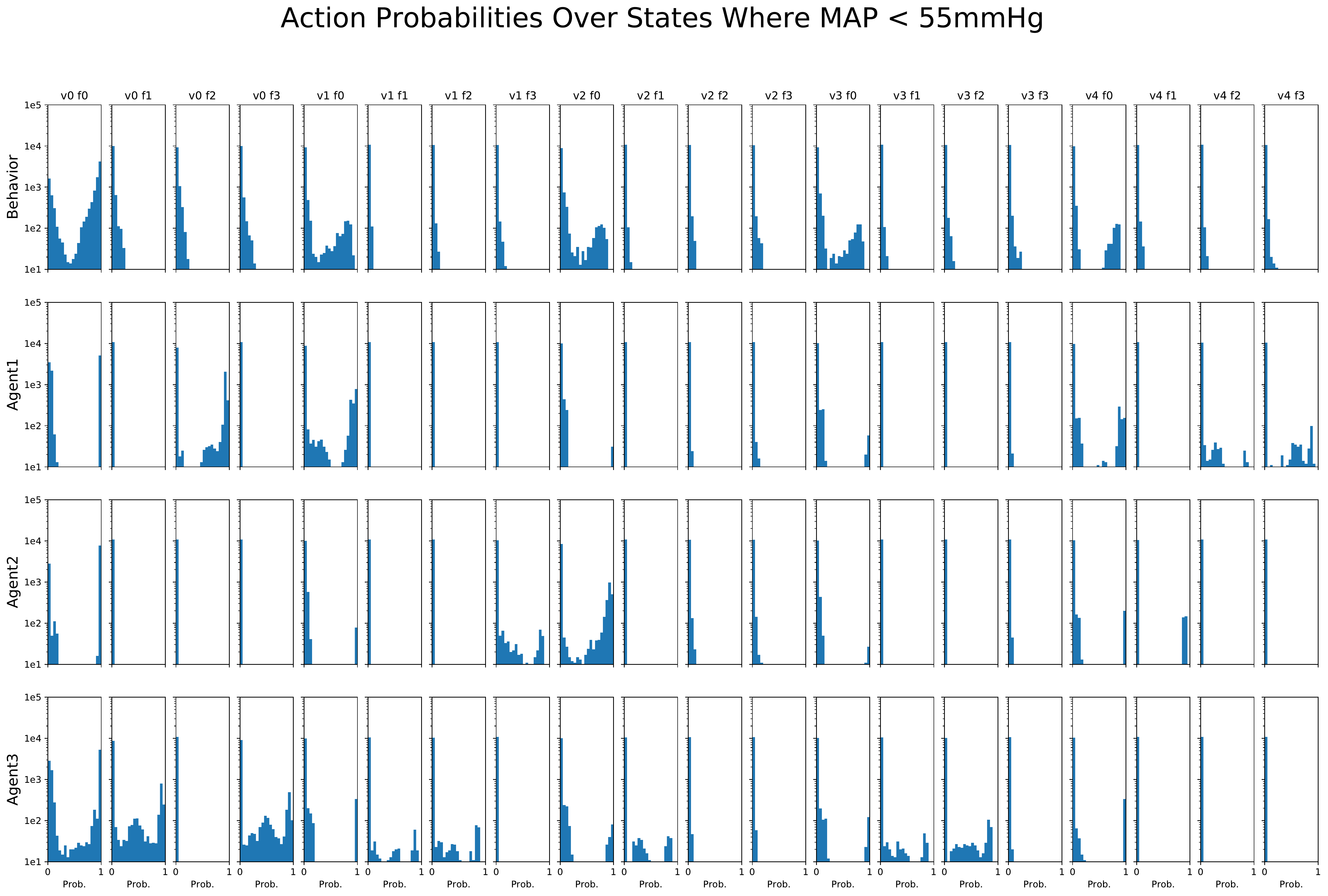}
 \end{center}
  \caption{Action probabilities for states with a low MAP of less than $55$mmHg. Each column corresponds to one of the 20 actions in our action space. The top row shows the physician behavior probabilities aggregated across all patients in the test set. The bottom three rows show probabilities from the three different agents, from the same run of the algorithm presented in the results table \ref{tab:Results}.}
 \label{fig:lowmap_actions_hist}
 \end{figure}

\clearpage
\newpage 
Finally, we show additional histograms of action probabilities for the 3 learned policies, along with the behavior policy, for several different subsets of states. We show how the behavior and learned policies focus on different actions in states where a fluid action was subsequently taken (Figure~\ref{fig:fluid_actions_hist}), states where a vasopressor action was subsequently taken (Figure~\ref{fig:vaso_actions_hist}), and states where the patient is in a stage of especially high acuity, as measured by elevated lactate (Figure~\ref{fig:highlactate_actions_hist}) and severely low MAP (Figure~\ref{fig:lowmap_actions_hist}).

\end{document}